%% file: main.tex
\documentclass[twoside,11pt]{article}

\usepackage{geometry}
\usepackage{crimson}
\usepackage{microtype}
\usepackage{booktabs}
\usepackage{url}
\usepackage{nicematrix} 
\usepackage{bm} 
\usepackage{eurosym} 
\usepackage{subcaption} 
\usepackage{siunitx} 
\usepackage{amsmath}
\usepackage{amsfonts}
\usepackage{mathtools}
\usepackage{amsthm}
\theoremstyle{plain}
\newtheorem{theorem}{Theorem}
\newtheorem{lemma}{Lemma}

\newtheorem{result}{Result}
\theoremstyle{definition}
\newtheorem{definition}{Definition}
\newtheorem{example}{Example}

\usepackage[colorlinks=false,allbordercolors={1 1 1}]{hyperref}
\hypersetup{hidelinks}
\usepackage[capitalize,noabbrev]{cleveref}
\usepackage[style=authoryear]{biblatex}
\bibliography{references}
\usepackage{tikz}
\usetikzlibrary{math} 
\usetikzlibrary{decorations.pathreplacing}
\usetikzlibrary{calligraphy}

\newcommand{\rv}{\bm}
\newcommand{\E}{\mathbb E}
\newcommand{\I}{\mathbb I}
\newcommand{\N}{\mathcal N}
\DeclareMathOperator{\Do}{do}
\DeclareMathOperator{\Var}{Var}
\DeclareMathOperator{\Bin}{Bin}
\DeclareMathOperator{\Dir}{Dir}

\DeclareMathOperator{\CB}{CB}
\DeclareMathOperator{\B}{B}
\newcommand{\dd}{\,\mathrm d}

\newcommand{\ceil}[1]{\left\lceil #1 \right\rceil}
\newcommand{\model}{\mathcal M}

\newcounter{save_theorem}
\newcommand{\repeatedthm}[4]{%
    \newcounter{#2_counter}%
    \setcounter{#2_counter}{\value{theorem}}%
    \begin{#1}%
        \label{#3}%
        #4%
    \end{#1}%
    \expandafter\newcommand\csname repeat#2\endcsname{%
        \setcounter{save_theorem}{\value{theorem}}
        \setcounter{theorem}{\value{#2_counter}}%
        \begin{theorem}%
            #4%
        \end{theorem}%
        \setcounter{theorem}{\value{save_theorem}}
    }%
}

\title{Uplift vs. predictive modeling: a theoretical analysis}

\author{%
  Théo Verhelst\textsuperscript 1,
  Robin Petit\textsuperscript 2,
  Wouter Verbeke\textsuperscript 3,
  Gianluca Bontempi\textsuperscript 1\\
  \textsuperscript 1Machine-learning Group
  Université Libre de Bruxelles, Belgium\\
  \textsuperscript 2Algorithms Research Group, Université Libre de Bruxelles, Belgium\\
  \textsuperscript 3Information Systems Engineering Research Group, KU Leuven, Belgium%
}

\date{}

\begin{document}

\maketitle

\begin{abstract}%
Despite the growing popularity of machine-learning techniques in decision-making, the added value of causal-oriented strategies with respect to pure machine-learning approaches has rarely been quantified in the literature. These strategies are crucial for practitioners in various domains, such as marketing, telecommunications, health care and finance. This paper presents a comprehensive treatment of the subject, starting from firm theoretical foundations and highlighting the parameters that influence the performance of the uplift and predictive approaches. The focus of the paper is on a binary outcome case and a binary action, and the paper presents a theoretical analysis of uplift modeling, comparing it with the classical predictive approach. The main research contributions of the paper include a new formulation of the measure of profit, a formal proof of the convergence of the uplift curve to the measure of profit ,and an illustration, through simulations, of the conditions under which predictive approaches still outperform uplift modeling. We show that the mutual information between the features and the outcome plays a significant role, along with the variance of the estimators, the distribution of the potential outcomes and the underlying costs and benefits of the treatment and the outcome.%
\end{abstract}

\paragraph{Keywords:}
Uplift modeling, Profit measure, Causal inference, Decision-making

\section{Introduction}
\label{sec:introduction}
With the growing popularity of machine-learning techniques in decision-making, the need for effective and accurate models has become increasingly important in various domains. Conventional predictive approaches have been used with success, for example, in churn prediction, where the models are built to forecast whether a customer is likely to stop using a service based on historical data~\parencite{oskarsdottir2018time,zhu2017empirical,mitrovic2018operational,idris2014ensemble}.

However, traditional predictive models often overlook an essential aspect of decision-making, the causal nature of interventions. Recently, uplift modeling has been established as an important approach to take this aspect into account for decision-making~\parencite{gutierrez2016causal,devriendt2021why}. Uplift modeling differs from conventional predictive models by explicitly considering the causal effect of an intervention on the outcome variable. Rather than estimating the conditional expectation of the outcome based on input features alone, uplift modeling focuses on estimating the difference in outcomes under different treatment scenarios.

Consider a marketing campaign for churn prevention, as an example. The goal is to identify customers who are less likely to churn in response to a promotional offer. Traditional predictive models predict the likelihood of customer churning, however, they do not consider the causal effect of the intervention (sending the offer) on the outcome (customer churn). In this setting, the possible behavior of a customer can be summarized in terms of counterfactual statements~\parencite{devriendt2021why}:
\begin{itemize}
    \item Sure thing: Customer not churning regardless of the action
    \item Persuadable: Customer churning only if not contacted
    \item Do-not-disturb: Customer churning only if contacted
    \item Lost cause: Customer churning regardless of the action
\end{itemize}
Ideally, only persuadable customers should be targeted by marketing actions. However, we observe only one of the two potential outcomes (this is known as the \emph{fundamental problem of causal inference}, \parencite{holland1986statistics}), and it is impossible to determine with certainty who are the persuadable customers. Uplift modeling explicitly aims to estimate the difference in the probability of a positive outcome under the treatment scenario (customer receives the offer) and the no-treatment scenario (customer does not receive the offer). Individuals maximizing this difference are the most likely to generate a profit increase when contacted. The term \emph{uplift} is used mainly in business settings where large amounts of experimental data are available, while in other fields, the same quantity is called the \emph{conditional average treatment effect} (CATE), or \emph{heterogeneous treatment effect}~\parencite{gutierrez2016causal}, usually assuming there is only access to observational data. A large number of models based on machine learning have been developed in recent years to estimate uplift, such as the S-learner, T-learner and X-learner~\parencite{zhang2021unified,kunzel2019metalearners}.

Despite the intuitive appeal of uplift modeling, the added value of causal-oriented strategies with respect to pure machine-learning predictive approaches has rarely been quantified in the literature. We believe that it is important to assess whether the expected benefit of uplift strategies (derived from a bias reduction in the estimation of causal effect) is still noticeable in settings where the data distribution is characterized by a large number of dimensions, nonlinearity, class imbalance and low class separability.

The works of \textcite{devriendt2021why,fernandez-loria2022causal,fernandez-loria2022causala,ascarza2018retention} address this issue. \textcite{devriendt2021why} and~\textcite{ascarza2018retention} present the uplift and predictive approaches, provide an empirical evaluation of both approaches and conclude that uplift models should be preferred over churn models. \textcite{fernandez-loria2022causal} develop an analytical criterion indicating when an uplift model leads to a lower causal classification error than a predictive model for a given individual. The same authors \parencite{fernandez-loria2022causala} discuss and develop the differences between causal classification and uplift modeling. and provide some qualitative arguments on when the predictive approach should be preferred. We extend these papers by comprehensively treating the question, starting from theoretical foundations and studying the influence of different characteristics of the setting (distribution of the outcome, variance of the estimators, etc.) on the performance of the uplift and predictive approaches.

A critical aspect of comparing the two approaches is the necessity for a meaningful and sensible measure of model performance. In this paper, we extend the work of ~\textcite{verbeke2021foundations} by developing a new formulation of the profit generated by a campaign where individuals targeted by interventions are selected by a machine-learning model. By incorporating the concept of profit, we go beyond the traditional evaluation metrics and consider the economic impact of decision-making strategies. Our measure of profit generalizes Verbeke's by accommodating varying costs and benefits across individuals. This flexibility is beneficial, for example, in churn prediction, where prioritizing higher-value customers is crucial. By selecting an appropriate measure, we ensure a fair and accurate comparison between the uplift and predictive models, enabling decision-makers to make informed choices based on the true effectiveness and suitability of each approach.

Our paper seeks to establish firm theoretical foundations for uplift modeling and to answer the question ``When does uplift modeling outperform predictive modeling?''. While we focus on a customer churn prediction example, our findings have broad applicability across domains, including marketing, telecommunications, health care and finance. Our main conclusions are as follows. The variance plays a critical role in determining the performance of a model, and in most cases, the predictive approach outperforms the uplift approach when the variance of the uplift estimator exceeds a certain threshold. We also show the important impact of three other aspects: cost sensitivity, the mutual information between the features and the outcome, and the distribution of the potential outcomes. While the importance of cost sensitivity and the distribution of potential outcomes have been discussed in the literature by~\textcite{verbeke2021foundations} and~\textcite{fernandez-loria2022causal}, respectively, to the best of our knowledge, the impact of mutual information has not been assessed before. We show that it has an important impact on performance, independent of the other aspects (estimator variance, cost sensitivity and distribution of potential outcomes).

Note, however, that we do not address the question of how to adapt uplift modeling to account for cost sensitivity or the other aspects mentioned above. Our contributions pertain to model \emph{evaluation} rather than model \emph{optimization}. Thus, it is left for future work to assess the effectiveness of cost-sensitive models in terms of the metrics developed in this paper. On that topic, \textcite{gubela2021uplift} have proposed a value-driven ranking method for targeted marketing campaigns.

The main research contributions of this paper are as follows:
\begin{itemize}
    \item A new formulation of the measure of profit, intensifying the focus on individual cost sensitivity and on the stochastic nature of the machine-learning model used to rank individuals (\cref{sec:campaign_profit}).
    \item A proof that the uplift curve (an evaluation curve often used in the uplift literature) is an estimator of the measure of profit, highlighting the strict conditions necessary for the validity of the uplift curve (\cref{sec:equivalence_uplift_curve}).
    \item An empirical estimator of the measure of profit, which is a cost-sensitive generalization of the uplift curve (\cref{sec:empirical_estimator}).
    \item A demonstration through theoretical analysis and simulations of the conditions under which the predictive approach still outperforms uplift modeling, and notably, the important role of the mutual information between the input features and the outcome, which has not previously been discussed in the literature (\cref{sec:uplift_predictive}).
\end{itemize}

The rest of this paper is organized as follows. \cref{sec:definitions} introduces the notations and notions used throughout this paper. Our contributions are presented in \cref{sec:profit_measure,sec:uplift_predictive}: we present the new formulation of the measure of profit in \cref{sec:profit_measure}, and we assess when the predictive approach outperforms the uplift approach in \cref{sec:uplift_predictive}. These results are further discussed in \cref{sec:discussion}. In \cref{sec:related_work}, we present related work on evaluation measures and on the comparison between the uplift and predictive approaches. Concluding remarks and recommendations for practitioners are given in \cref{sec:conclusion}. Proofs of the theorems are provided in \cref{sec:appendix_equivalence,sec:appendix_properties_simulation}.

\section{Background}
\label{sec:definitions}
In this section, we introduce the notations and present the key concepts used throughout this paper.

\subsection{Notation}
\label{sec:notation}
\begin{table}
    \begin{center}
    \begin{minipage}{0.95\textwidth}
    \caption{Mathematical notation.}\label{tab:notation}%
    \begin{tabular}{ll}
        \toprule
        Notation & Definition \\
        \midrule
        $\rv v$ & Random variable \\
        $v$ & Realization of $\rv v$ \\
        $\rv y\in\{0,1\}$ & Outcome indicator \\
        $\rv t\in\{0,1\}$ & Treatment indicator \\
        $\rv x\in\mathcal X\subseteq\mathbb R^n$ & Set of features \\
        $f_{\rv x}(x)$ & Probability density function of $\rv x$ \\
        $\mathbb I[\cdot]$ & Iverson bracket, equals one when the expression between\\
        &\hspace{1em}brackets is true, zero otherwise \\
        $\Do(\rv t=t)$ & Causal intervention $\rv t=t$ \\
        $S_t=P(\rv y_t=y)$ & Probability of the outcome $\rv y=y$ under $\Do(\rv t=t)$ \\
        $S_0(x), S_1(x)$ & $P(\rv y_0=1\mid \rv x=x),P(\rv y_1=1\mid \rv x=x)$ \\
        $U$ & Uplift, defined as $U=S_0-S_1$ \\
        $\rv D=\{(\rv x^{(i)}, \rv y^{(i)}, \rv t^{(i)})\}_{i=1}^N$ & Training set or test set \\
        $\model(\rv x, \rv D_{\mathrm{tr}})$ & Prediction for features $\rv x$ of model $\model$ trained on set $\rv D$ \\
        $\tau\in\mathbb R$ & Classification threshold \\
        $\rho\in[0,1]$ & Treatment rate \\
        \bottomrule
    \end{tabular}
    \end{minipage}
    \end{center}
\end{table}

We use Pearl's causal framework, which is based on the notion of \emph{structural causal models} (SCM). A formal definition of SCMs is given by~\textcite[Def. 7.1.1]{pearl2009causality}. Here, $\rv t$ is a random variable denoting the \emph{action}, or \emph{treatment}, $\rv y$ is the \emph{outcome}, and $\rv x$ is a set of \emph{features} (or \emph{covariates}) describing the unit/individual. We denote the realizations of these variables as $\mathcal \rv t, \mathcal \rv y$ and $\mathcal \rv x$, respectively. In this paper, we will limit ourselves to considering the double binary causal classification case, that is, the setting where $\rv y\in\{0,1\}$ and $\rv t\in\{0,1\}$. Importantly, we always assume having access to experimental data, in which the treatment $\rv t$ is randomized. It is possible to learn the uplift from observational data, for example, with propensity scores \parencite{kunzel2019metalearners,curth2021nonparametric} or double machine learning~\parencite{jung2021estimating}, however, this is beyond the scope of this paper. The $\Do(\rv t=t)$ operator denotes a causal intervention in the system. The conditional probability of $\rv y=y$ given $\rv x=x$ under intervention $\Do (\rv t=t)$ is written as $P(\rv y = y\mid \Do(\rv t=t), \rv x=x)$, or $P(\rv y_t=y\mid \rv x=x)$. For ease of notation, we also define
\begin{align}
    S_0(x)&=P(\rv y_0=1\mid \rv x=x) & S_0 &= P(\rv y_0=1) \\
    S_1(x)&=P(\rv y_1=1\mid \rv x=x) & S_1 &= P(\rv y_1=1) \\
    U(x) &= S_0(x)-S_1(x)    & U   &= S_0 - S_1.
\end{align}
In this notation, $U$ is the \emph{uplift}, or \emph{average treatment effect} (ATE), and $U(x)$ is the \emph{individual uplift}, or \emph{conditional average treatment effect} (CATE). Note that, for example, in the literature pertaining to retail or online advertisements, the uplift is defined as $U=S_1-S_0$, and similarly $U(x)=S_1(x)-S_0(x)$. This choice depends on whether the probability of the (positive) outcome $\rv y=1$ should be minimized (e.g., in churn prevention) or maximized (e.g., in sales). The uplift is then defined so that a positive uplift corresponds to a beneficial outcome. Since we apply our results mostly to churn prevention, we use the convention $U=S_0-S_1$.

Let $\model$ be a model that is used to rank individuals such that only the individuals with the highest scores should be targeted by the action. The model $\model$ is trained from a data set $D=\{(x^{(i)},y^{(i)},t^{(i)})\}_{i=1}^N$ of $N$ iid\footnote{The independence assumption might be violated in applications such as churn with, for example, a word-of-mouth effect generating a second order of treatment.} realizations of $(\rv x, \rv y, \rv t)$. We assume that $D$ is the result of a random process, and we denote it as a random variable as $\rv D$. We consider $\model(x, D_{\mathrm{tr}})$ as a learning algorithm, taking a data set $D$ and a set of features $\rv x$ as input and returning a score for $x$, for example, an estimation of $U(x)$.

A threshold $\tau$ is used to determine which individuals should be targeted. The model $\model$ prescribes targeting all individuals with a score $\model(x, D_{\mathrm{tr}})\ge\tau$ and not targeting the remaining individuals. The threshold $\tau$ depends upon the model being used, because different models can provide scores in different ranges, and that are differently distributed. Therefore, to consistently compare the performance of different models, we let $\rho\in(0,1)$ be the proportion of individuals who should be targeted, and the corresponding threshold $\tau$ can be determined as the largest value that satisfies $\rho=P(\model(\rv x, D_{\mathrm{tr}})>\tau)$. Note the random variable $\rv x$ in this expression. Since $\model(x, D_{\mathrm{tr}})$ is a deterministic function of $x$ (for a given $D$), $\model(\rv x, D_{\mathrm{tr}})$ is a random variable for which we can compute the probability $P(\model(\rv x, D_{\mathrm{tr}})>\tau)$.

\subsection{Uplift and predictive approaches}
In this paper, we designate by the \emph{predictive approach} a machine-learning model $\mathcal M_p$ that estimates the conditional probability $S_0(x)=P(\rv y_0=1\mid\rv x=x)$:
\begin{equation}
    \mathcal M_p(x,D_{\mathrm{tr}})\approx S_0(x).
\end{equation}
We designate by the \emph{uplift approach} a machine-learning model $\mathcal M_u$ that estimates the uplift $U(x)=P(\rv y_0=1\mid\rv x=x)-P(\rv y_1=1\mid\rv x=x)$:
\begin{equation}
    \mathcal M_u(x,D_{\mathrm{tr}})\approx U(x).
\end{equation}
Note that the definition of these approaches can vary in the literature. \textcite{fernandez-loria2022causal} focus on online advertisements, in which the outcome should be maximized. As such, they define three approaches; the \emph{treatment difference} (TD) approach, which is defined as $S_1(x)-S_0(x)$ (the opposite of the uplift approach as defined in this paper); the \emph{outcome most} (OM) approach, which ranks individuals by $S_1(x)$; and the \emph{outcome least} approach (OL), which ranks individuals by $1-S_0(x)$. The TD and OM approaches are equivalent to our uplift and predictive approaches, respectively, up to a change in label for the values of $\rv t$.

\subsection{Uplift curve}
\label{sec:uplift_curve}
The uplift curve quantifies the quality of the ranking provided by a model $\model$ as a function of the proportion of selected individuals. It is used in most papers in the uplift literature~\parencite{gutierrez2016causal}. The model $\model$ is used to predict a score for every sample of a test data set $\rv D_{\mathrm{te}}=\{(\rv x^{(i)}, \rv y^{(i)}, \rv t^{(i)})\}_{i=1}^N$. The uplift is then estimated by comparing the outcome rate in the control and target groups among the individuals with the highest scores. \textcite{devriendt2020learning} compare the different definitions of the uplift curve and the related Qini curve in the literature. In their framework, we focus on the absolute, joint definition of the uplift curve (Equation (8) in their paper).
\begin{definition}
    \label{def:simple_uplift_curve}
    Let $\rv D_{\mathrm{te}}=\{(\rv x^{(i)},\rv y^{(i)},\rv t^{(i)})\}_{i=1}^N$ be a data set of $N$ iid tuples of random variables, where the treatment $\rv t^{(i)}$ is randomized. Let $\model$ be a model trained on a data set $D_{\mathrm{tr}}$, and let $\rv D_{\mathrm{te}}$ be sorted in decreasing order according to $\model$: for any $i<j$, we have $\model(\rv x^{(i)}, D_{\mathrm{tr}})\ge\model(\rv x^{(j)}, D_{\mathrm{tr}})$. The \emph{uplift curve} is defined for $k\in\{1,\dots,N\}$ as
    \begin{equation}
        \label{eq:simple_def_uplift_curve}
        \mathrm{Uplift}(k,D_{\mathrm{tr}},\rv D_{\mathrm{te}})=\left(\frac{\rv r_0(k)}{\rv n_0(k)}-\frac{\rv r_1(k)}{\rv n_1(k)}\right)k
    \end{equation}
    where the following notation is used for $t=0,1$:
    \begin{align*}
        \rv r_t(k) &= \sum_{i=1}^k\I[\rv y^{(i)}=1\text{ and }\rv t^{(i)}=t], & \rv n_t(k) &= \sum_{i=1}^k\I[\rv t^{(i)}=t].
    \end{align*}
    In the case where $\rv r_t(k)=\rv n_t(k)=0$, the quotient $\rv r_t(k)/\rv n_t(k)$ is defined as $0$.
\end{definition}
In this definition, the quantity $\rv n_t(k)$ represents the number of individuals with treatment $\rv t=t$ among the $k$ individuals with the highest scores. The quantity $\rv r_t(k)$ represents the number of individuals who responded (i.e., such that $\rv y=1$) with treatment $\rv t=t$ among the $k$ individuals with the highest scores.

\subsection{Verbeke's measure of profit}
\label{sec:verbeke_definition}
As illustrated by~\textcite{gubela2021uplift}, using the uplift approach for ranking individuals and using the associated uplift curve for evaluating a model does not take into account the cost and benefits associated with each individual and with the action $\rv t=1$. We present in \cref{sec:related_work} an overview of the different papers addressing this issue. The most general measure, in our opinion, for evaluating the performance of uplift models was proposed by~\textcite{verbeke2022not}. Its generality stems from the fact that it is not tied to a specific operational setting (e.g., churn prediction or online retail). This is achieved by defining a cost-benefit matrix (see below) that effectively captures the diverse range of settings characterized by cost sensitivity. The evaluation measure that we present in \cref{sec:profit_measure} is mostly equivalent to Verbeke's (this equivalence is proven in \cref{sec:equivalence_verbeke}). Our measure more clearly emphasizes the influence of the machine-learning model and the individual costs and benefits in the population. In \cref{sec:uplift_predictive}, we employ these differences to assess the factors that impact the performance of the uplift and predictive approaches.

Let $F_{yt}^{D_{\mathrm{tr}}}$ be the cumulative distribution function of the score from a model $\model$ trained on a data set $D_{\mathrm{tr}}$, conditional on a particular realization of the potential outcome $\rv y_t=y$:
\begin{equation*}
    F_{yt}^{D_{\mathrm{tr}}}(\tau)=P(\model(\rv x, D_{\mathrm{tr}})<\tau\mid \rv y_t=y).
\end{equation*}
Now, let the causal confusion matrix $\mathrm{CF}(\tau, D_{\mathrm{tr}})$ for a threshold $\tau$ be
\begin{equation*}
    \mathrm{CF}(\tau, D_{\mathrm{tr}})=\begin{bNiceMatrix}[first-row,last-col]
            \rv t=0    & \rv t=1    &     \\
            (1-S_0)F^{D_{\mathrm{tr}}}_{00}(\tau) & (1-S_1)(1-F^{D_{\mathrm{tr}}}_{01}(\tau)) & \rv y=0 \\
            S_0F^{D_{\mathrm{tr}}}_{10}(\tau) & S_1(1-F^{D_{\mathrm{tr}}}_{11}(\tau))   & \rv y=1
        \end{bNiceMatrix}.
\end{equation*}
Then, let $\mathrm E$ be the \emph{causal effect matrix}, defined as
\begin{equation*}
    \mathrm E(\tau,D_{\mathrm{tr}})=\mathrm{CF}(\tau,D_{\mathrm{tr}})-\mathrm{CF}(\infty,D_{\mathrm{tr}})
\end{equation*}
with $\mathrm{CF}(\infty,D_{\mathrm{tr}})$ defined as
\begin{equation*}
    \mathrm{CF}(\infty,D_{\mathrm{tr}})=\begin{bmatrix}
        1-S_0 & 0 \\ S_0 & 0
    \end{bmatrix}.
\end{equation*}
Finally, we define a cost-benefit matrix $\CB$ that expresses the sum of the costs and benefits for the two possible actions ($\rv t=0$ or $\rv t=1$) and the two possible outcomes ($\rv y=0$ or $\rv y=1$). In this definition, the cost-benefit matrix is the same for all the individuals. It is noted
\begin{equation}
    \label{eq:cb_matrix}
    \CB=\begin{bNiceMatrix}[first-row,last-col]
        \rv t=0         & \rv t=1         &     \\
        \CB_{00} & \CB_{01} & \rv y=0 \\
        \CB_{10} & \CB_{11} & \rv y=1
    \end{bNiceMatrix}
\end{equation}
For ease of notation, we note the sum of the elements in the componentwise product of the matrices $A$ and $B$ (also called the Frobenius inner product) as  $A\oplus B$:
\begin{equation*}
    A\oplus B=\sum_{ij}A_{ij}B_{ij}.
\end{equation*}
This operation satisfies the same axioms as the inner product.
\begin{definition}[\parencite{verbeke2022not}]
    \label{def:verbeke_profit_measure}
    The measure of causal profit $\mathrm{CP}(\tau, D_{\mathrm{tr}})$, for a threshold $\tau$, a training set $D$ and a constant cost-benefit matrix $\CB$, is defined as
    \begin{equation}
        \mathrm{CP}(\tau, D_{\mathrm{tr}})=\mathrm E(\tau, D_{\mathrm{tr}})\oplus \CB.
    \end{equation}
\end{definition}
We will show in \cref{sec:equivalence_verbeke} that with the exception of minor technical differences, this evaluation measure is equivalent to the one we develop in our paper.

\section{Measure of profit}
\label{sec:profit_measure}
This section presents our research contributions to the performance evaluation of a model in a causal decision-making setting. We start by introducing the concept of causal profit for individuals, which measures the incremental profit gained by targeting specific individuals with interventions. Next, we extend this definition to the campaign level, where the cumulative profit is assessed by considering the overall impact of targeting a group of individuals. To establish the connection between the causal profit and existing measures, we first prove its equivalence with Verbeke's original definition of profit~\parencite{verbeke2021foundations}. Then, we prove that the uplift curve is an estimator of the causal profit under a specific assumption on the values of the costs and benefits generated by the individuals. Finally, we propose an empirical estimator of the profit measure that leverages the data, and a ranking model to estimate the potential profitability of targeting specific individuals. This new empirical performance metric is a cost-sensitive generalization of the uplift curve.

\subsection{Individual profit}
\label{sec:individual_profit}
Our measure of profit for a campaign, which is presented in the next section, relies on the notion of \emph{individual causal profit}. This notion was already defined by \textcite{gubela2021uplift} and \textcite{gubela2020response}, which they name \emph{revenue uplift}. It is a generalization of the individual uplift $U(x)$ to cost-sensitive settings. Our contribution to individual causal profit with respect to the definition provided by \textcite{gubela2021uplift} and \textcite{gubela2020response} lies in the clear separation between the cost-benefit matrix and the outcome variable $\rv y$. First, we need to define the cost-benefit matrix at the individual level $\rv x=x$. This generalization of $\CB$ to the individual level is useful in settings such as churn prediction, where different customers have different values (e.g., some customers might have a more expensive tariff plan than others); hence, to maximize profits, the retention efforts should be focused on high-value customers.
\begin{definition}
    \label{def:cb_x_matrix}
    The cost-benefit matrix $\CB(x)$ expresses the sum of the costs and benefits for the two possible actions ($\rv t=0$ or $\rv t=1$) and the two possible outcomes ($\rv y=0$ or $\rv y=1$) for an individual with features $\rv x=x$. It is noted
    \begin{equation}
        \label{eq:cb_x_matrix}
        \CB(x)=\begin{bNiceMatrix}[first-row,last-col]
            \rv t=0         & \rv t=1         &     \\
            \CB_{00}(x) & \CB_{01}(x) & \rv y=0 \\
            \CB_{10}(x) & \CB_{11}(x) & \rv y=1
        \end{bNiceMatrix}
    \end{equation}
\end{definition}
Note that, although the actual value generated by an individual is inherently a random variable (e.g., the data consumption of a customer is not known in advance), $\CB(x)$ expresses the expected cost-benefits for all individuals with features $\rv x=x$. From this matrix, and the probability distribution of the outcome $\rv y$, one can define the expected profit that a given action generates:
\begin{definition}
    \label{def:ind_profit}
    When action $\rv t=t$ is carried out for an individual $\rv x=x$, we define the \emph{individual profit} of that action as
    \begin{equation}
        \label{eq:ind_profit}
        \pi_t(x)=\CB_{0t}(x)(1-S_t(x))+\CB_{1t}(x)S_t(x).
    \end{equation}
\end{definition}
As demonstrated by~\textcite{verbeke2021foundations}, the performance of a model should be measured relative to a baseline scenario rather than in absolute terms. That is because, even when no action is carried out, an outcome will always occur, and therefore, the success of an action should be compared with the outcome resulting from the absence of action. Formally, we name this difference the \emph{causal profit}:

\begin{definition}
    \label{def:ind_causal_profit}
    The \emph{individual causal profit} for an individual with features $\rv x=x$ is defined as
    \begin{align}
        \label{eq:ind_causal_profit}
        \pi(x)&=\pi_1(x)-\pi_0(x).
    \end{align}
\end{definition}
We can readily develop this expression to obtain
\begin{align}
    \pi(x)=&\CB_{01}(x)(1-S_1(x))+\CB_{11}(x)S_1(x)-\CB_{00}(x)(1-S_0(x)) - \CB_{10}(x)S_0(x)\label{eq:pi_cb}
\end{align}
We can see that the causal profit is a function of $S_0(x)$, $S_1(x)$ and $\CB(x)$. This quantity was defined similarly by \textcite[Eq. 42]{verbeke2021foundations}. We now illustrate these definitions in the churn prevention use case, with marketing campaigns.
\begin{example}
    Let us suppose that our objective is to prevent customer churn with marketing campaigns. The costs and benefits of this campaign are defined as follows:
    \begin{itemize}
        \item Calling a customer has an operating cost $C$
        \item We gain value $V_0$ when the customer does not churn (a value called the \emph{customer lifetime value}), and we gain $V_1$ when they churn (typically $\text{\euro}0$)
        \item We offer a marketing incentive of cost $I$ to the customer when we call them, and if they do not churn
    \end{itemize}
    From this, we can compute the cost-benefit matrix $\CB(x)$ as
    \begin{equation*}
        \CB(x)=\begin{bNiceMatrix}[first-row,last-col]
            \rv t=0    & \rv t=1    &     \\
            V_0 & V_0-C-I & \rv y=0 \\
            V_1 & V_1-C   & \rv y=1
        \end{bNiceMatrix}
    \end{equation*}
    and the causal profit $\pi(x)$ is
    \begin{align*}
        \pi(x)&=-C-I-(V_0-I-V_1)S_1(x)+(V_0-V_1)S_0(x)\\
        &=(V_0-V_1)U(x)-C-IP(\rv y_1=0\mid x).
    \end{align*}
    Hence, if the magnitude of the uplift and the expected lifetime value of the customer are greater than the expected cost of the retention action, then we can expect a positive causal profit by calling the customer. To demonstrate, let us suppose that the customer lifetime value is $V_0=\text{\euro}120$, $V_1=\text{\euro}0$, the call has an operating cost $C=\text{\euro}1$, the marketing incentive has a cost $I=\text{\euro}20$, and this customer has churn probabilities of $S_0(x)=0.15$ and $S_1(x)=0.05$, hence an uplift of $U(x)=0.1$. The cost-benefit matrix is
    \begin{equation*}
        \CB(x)=\begin{bNiceMatrix}[first-row,last-col]
            \rv t=0    & \rv t=1    &     \\
            120 & 99 & \rv y=0 \\
            0   & -1   & \rv y=1
        \end{bNiceMatrix}
    \end{equation*}
    and the causal profit is $\pi(x)=120\times 0.1-1-20\times 0.95=-8$. Given the low uplift of this customer and the high cost of the incentive, they should therefore not be targeted by this marketing campaign.
\end{example}

\subsection{Campaign profit}
\label{sec:campaign_profit}
The causal profit $\pi(x)$, as per \cref{def:ind_causal_profit}, is defined per individual $\rv x=x$, but in business applications, a campaign is carried out on a large number of individuals. As presented in \cref{sec:verbeke_definition}, \textcite{verbeke2022not} define a cost-sensitive measure of the causal profit of a campaign based on the causal confusion matrix and the cost-benefit matrix. In this section, we provide another definition of the causal profit of a campaign, which we then show to be equivalent to the definition given by Verbeke et al. Our contribution lies in the fact that our measure emphasizes the individual causal profit, and highlights the separate influence of the individual cost, benefits and the stochastic nature of the machine-learning estimator. We also give a formal proof that the uplift curve, which is widely used in the uplift literature, is equivalent to the measure of profit. In particular, we highlight the strict assumption necessary for the validity of the uplift curve.

As defined in \cref{sec:notation}, let us assume that action $\Do(\rv t=1)$ is carried out on the proportion $\rho$ of individuals with the highest scores, and action $\Do(\rv t=0)$ is carried out on the other individuals. We call $\rho$ the \emph{treatment rate}. If we have a population of $N$ individuals, then $\ceil{N\rho}$ individuals will be targeted. Then, given a predictive model $\model(\rv x, D_{\mathrm{tr}})$, we can find the threshold $\tau$ on the scores that would separate the proportion $\rho$ of individuals with the highest scores from the rest. Formally, $\tau$ is defined as
\begin{equation}
    \tau=\inf\,\{\tau':P(\model(\rv x, D_{\mathrm{tr}})\ge\tau')=\rho\}.
\end{equation}
As for the individual profit defined in \cref{sec:individual_profit}, it is important to note that even if a campaign is not carried out, some profit will be generated anyway. Therefore, the performance of an uplift model should be evaluated in terms of profit with respect to a baseline scenario where no action is taken. Here, we first define the profit induced by carrying out the campaign (\cref{def:action_profit}) and the profit of the baseline scenario where the campaign is not carried out (\cref{def:baseline_profit}). Finally, the causal profit is defined as the difference between these two quantities (\cref{def:causal_profit}).
\begin{definition}
    \label{def:action_profit}
    The \emph{action profit} of a campaign with a treatment rate $\rho$ and a model $\model$ trained on a data set $D_{\mathrm{tr}}$ is defined as
    \begin{align}
        \label{eq:action_profit}
        \Pi_1(\rho,D_{\mathrm{tr}})&=\rho\E_{\rv x}[\pi_1(\rv x)\mid \model(\rv x, D_{\mathrm{tr}})\ge\tau]+(1-\rho)\E_{\rv x}[\pi_0(\rv x)\mid \model(\rv x, D_{\mathrm{tr}})<\tau]
    \end{align}
    where $\tau$ is defined as $\tau=\inf\,\{\tau':P(\model(\rv x, D_{\mathrm{tr}})\ge\tau')=\rho\}$.
\end{definition}
Intuitively, this quantity is the sum of the expected profit of the group of individuals being targeted (i.e., those with a score $\model(\rv x, D_{\mathrm{tr}})\ge\tau$) and the expected profit of the group of individuals not being targeted (i.e., those with a score $\model(\rv x, D_{\mathrm{tr}})<\tau$). It is important to note that the quantity in \cref{def:action_profit}, as well as in the following definitions, is independent of the size of the population. As such, \cref{eq:action_profit} represents the expected profit generated, on average, per individual in. As a very simple example, if we have a campaign on a population where targeted individuals generate a profit of $\text{\euro}20$, and nontargeted individuals generate $\text{\euro}10$, a treatment rate of $\rho=0.5$ would result in an action profit $\Pi_1(\rho, D_{\mathrm{tr}})=\text{\euro}15$.

Now, we define the baseline profit of the campaign as follows.
\begin{definition}
    \label{def:baseline_profit}
    The \emph{baseline profit} of running no campaign is defined as
    \begin{equation}
        \label{eq:baseline_profit}
        \Pi_0=\E_{\rv x}[\pi_0(\rv x)].
    \end{equation}
\end{definition}
The causal profit of the campaign is logically defined as the difference between these two quantities.
\begin{definition}
    \label{def:causal_profit}
    The \emph{causal profit} of a campaign with a treatment rate $\rho$ and a model $\model$ trained on a data set $D_{\mathrm{tr}}$ is defined as
    \begin{align}
        \label{eq:causal_profit}
        \Pi(\rho,D_{\mathrm{tr}})&=\Pi_1(\rho,D_{\mathrm{tr}})-\Pi_0.
    \end{align}
\end{definition}
Now, recall that the training set $D_{\mathrm{tr}}$ is fixed in the previous definitions. Let us take into account the stochastic nature of the sampling process of $\rv D_{\mathrm{tr}}$, which leads the model $\model$ to provide different scores depending on the realization of the training set. This generalization is necessary if one wants to formalize the variance and bias of the estimator $\model$. To the best of our knowledge, none of the evaluation measures for uplift modeling in the literature take into account the random aspect of $\rv D_{\mathrm{tr}}$. This generalization is essential to the comparison of the uplift and predictive approaches in \cref{sec:uplift_predictive}. Concretely, we define the expected causal profit as the expected value of the causal profit over the distribution of $\rv D_{\mathrm{tr}}$.
 \begin{definition}
    \label{def:expected_causal_profit}
    The \emph{expected causal profit} of a campaign with a treatment rate $\rho$ and a model $\model$ trained on a distribution of data sets $\rv D$ is defined as
    \begin{align}
        \overline\Pi(\rho)=\E_{\rv D_{\mathrm{tr}}}[\Pi(\rho,\rv D_{\mathrm{tr}})].
    \end{align}
\end{definition}
These various definitions can be unwrapped and expressed in terms of just $\pi(x)$ and the score $\model(x, D_{\mathrm{tr}})$ over the distribution of $\rv x$. In particular, the causal profit can be computed only from the causal profit of only the targeted individuals. This is formalized in the following theorem.
\begin{theorem}
\label{thm:total_profit}
Let the model $\model(\rv x, D_{\mathrm{tr}})$ be a continuous random variable for all realizations $D_{\mathrm{tr}}$ of $\rv D_{\mathrm{tr}}$. The causal profit can be expressed as
    \begin{equation}
        \label{eq:thm_profit}
        \Pi(\rho,D_{\mathrm{tr}})=\E_{\rv x}[\pi(\rv x)\I[\model(\rv x, D_{\mathrm{tr}})\ge\tau]]
    \end{equation}
    where $\tau$ is a constant defined as $\tau=\inf\,\{\tau':P(\model(\rv x, D_{\mathrm{tr}})\ge\tau')=\rho)\}$,
    and the expected causal profit can be expressed as\footnote{Note that in \cref{eq:thm_profit_total}, the probability $P(\model(\rv x, \rv D_{\mathrm{tr}})\ge\tau)$ is taken over the distribution of training sets $\rv D$, and not on $\rv x$ as a random variable, because the variable $\rv x$ is bound by the surrounding expected value. This probability is equal to $\E_{\rv D}[\I[\model(\rv x, \rv D_{\mathrm{tr}})\ge\tau]]$.}
    \begin{equation}
        \label{eq:thm_profit_total}
        \overline\Pi(\rho)=\E_{\rv x}[\pi(\rv x)P(\model(\rv x, \rv D_{\mathrm{tr}})\ge\tau)]
    \end{equation}
    where $\tau$ is a random variable\footnote{In \cref{eq:thm_profit}, $D$ is a constant and thus $\tau$ is a constant as well, while in \cref{eq:thm_profit_total} $\rv, D$ is a random variable and thus $\tau$ is a random variable as well.} defined as $\tau=\inf\,\{\tau':P(\model(\rv x, \rv D_{\mathrm{tr}})\ge\tau')=\rho)\}$.
\end{theorem}
\begin{proof}
    Let $f_{\rv x}(\cdot)$ be the probability density function of $\rv x$. We can develop the conditional expectations in $\Pi_1(\rho,D_{\mathrm{tr}})$ as
    \begin{align}
        \Pi_1(\rho,D_{\mathrm{tr}})=&\frac\rho{P(\model(\rv x, D_{\mathrm{tr}})\ge\tau)}\int f_{\rv x}(x)\pi_1(x)\I[\model(x, D_{\mathrm{tr}})\ge\tau]\dd x \nonumber\\
        &+\frac{1-\rho}{{P(\model(\rv x, D_{\mathrm{tr}})<\tau)}}\int f_{\rv x}(x)\pi_0(x)\I[\model(x, D_{\mathrm{tr}})<\tau]\dd x.\label{eq:proof_thm_profit_total}
    \end{align}
    $\model(\rv x, D_{\mathrm{tr}})$ is a continuous random variable; therefore, its cumulative distribution function is monotonically increasing. Hence, the value $\tau$ that satisfies its definition $\tau=\inf\,\{\tau':P(\model(\rv x, D_{\mathrm{tr}})\ge\tau')\ge\rho\}$ in fact satisfies exactly $\rho=P(\model(\rv x, D_{\mathrm{tr}})\ge\tau)$. Hence, \cref{eq:proof_thm_profit_total} simplifies to
    \begin{align*}
        =&\int f_{\rv x}(x)\pi_1(x)\I[\model(x, D_{\mathrm{tr}})\ge\tau]\dd x+\int f_{\rv x}(x)\pi_0(x)\I[\model(x, D_{\mathrm{tr}})<\tau]\dd x\\
        =&\int f_{\rv x}(x)(\pi_1(x)\I[\model(x, D_{\mathrm{tr}})\ge\tau]+\pi_0(x)(1-\I[\model(x, D_{\mathrm{tr}})\ge\tau]))\dd x\\
        =&\int f_{\rv x}(x)(\pi_0(x)+\pi(x)\I[\model(x, D_{\mathrm{tr}})\ge\tau])\dd x\\
        =&\Pi_0 + \E_{\rv x}[\pi(\rv x)\I[\model(\rv x, D_{\mathrm{tr}})\ge\tau]].
    \end{align*}
    Therefore, the causal profit of the campaign can be expressed as
    \begin{align*}
        \Pi(\rho,D_{\mathrm{tr}})=\Pi_1(\rho,D_{\mathrm{tr}})-\Pi_0= \E_{\rv x}[\pi(\rv x)\I[\model(\rv x, D_{\mathrm{tr}})\ge\tau]].
    \end{align*}
    The expected causal profit can be developed as
    \begin{align*}
        \overline\Pi(\rho)&=\E_{\rv D}[\Pi(\rho,\rv D_{\mathrm{tr}})]\\
        &=\E_{\rv D}[\E_{\rv x}[\pi(\rv x)\I[\model(\rv x, D_{\mathrm{tr}})\ge\tau]]]\\
        &=\E_{\rv x}[\pi(\rv x)\E_{\rv D}[\I[\model(\rv x, D_{\mathrm{tr}})\ge\tau]]]\\
        &=\E_{\rv x}[\pi(\rv x)P(\model(\rv x, \rv D_{\mathrm{tr}})\ge\tau)].
    \end{align*}
\end{proof}

Let us illustrate these definitions and \cref{thm:total_profit} with a simplistic example.
\begin{example}
    \label{ex:profit}
    Let us suppose that we have a population of six individuals with feature values $x^{(1)},\dots,x^{(6)}$. The expected profits are given in \cref{tab:example_profits}, and the resulting ranking from a model $\model$ is depicted in \cref{fig:example_campaign_profit}. Note that the ranking provided by $\model$ is based here on the expected profit, however, in general, this is not always the case.

    \begin{table}[]
        \centering
        \caption{Expected profits of the population of six individuals from \cref{ex:profit}.}
        \label{tab:example_profits}
        \begin{tabular}{cSSSSSS}
        \toprule
        & $x^{(1)}$ & $x^{(2)}$ & $x^{(3)}$ & $x^{(4)}$ & $x^{(5)}$ & $x^{(6)}$ \\
        \midrule
         $\pi_0(x)$ & -0.1 & 0.1 & 0.15 & 0.1 & 0.2 & 0 \\
         $\pi_1(x)$ &  0.2 & 0.05 & -0.05 & 0.1 & -0.1 & 0.1 \\
         $\pi(x)$ & 0.3 & -0.05 & -0.2 & 0 & -0.3 & 0.1 \\
         \bottomrule
        \end{tabular}
    \end{table}
    
    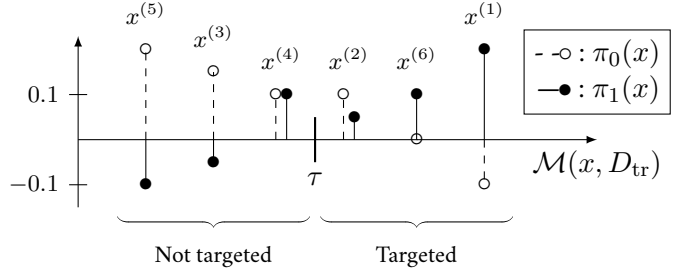
\begin{figure}
        \centering
        \tikzmath{
            \incr=0.15;
            \gap=0.012;
        }
        \begin{tikzpicture}[
            x=6cm, y=6cm,
            brace/.style={
                decorate,
                decoration={
                    calligraphic brace,
                    raise=-5pt,
                    amplitude=4pt
        }}]
        \draw[-latex, thin] (0, 0)--(1.15, 0) node[below] {$\model(x, D_{\mathrm{tr}})$};
        \draw[-latex, thin] (0, -0.15)--(0, 0.23);
        \draw (1.15, 0.15) node [draw,align=left] {- -\hspace{-0.1em}$\circ$ : $\pi_0(x)$\\\hspace{0.1em}---\hspace{-0.1em}$\bullet$ : $\pi_1(x)$};
    
        \draw (-0.02, 0.1) node[left] {\footnotesize $0.1$} -- (0.02, 0.1);
        \draw (-0.02, -0.1) node[left] {\footnotesize $-0.1$} -- (0.02, -0.1);

        \draw[thick] (3.5*\incr, -0.05) node[below] {$\tau$} -- (3.5*\incr, 0.05);
    
        \foreach \i/\j/\pzero/\pone/\thegap in {
            1/5/0.2/-0.1/0, 2/3/0.15/-0.05/0, 3/4/0.1/0.1/\gap,
            4/2/0.1/0.05/\gap, 5/6/0/0.1/0, 6/1/-0.1/0.2/0
        } {
            \draw[dashed] (\i*\incr-\thegap, 0) -- (\i*\incr-\thegap, \pzero) node {$\circ$};
            \draw (\i*\incr+\thegap, 0) -- (\i*\incr+\thegap, \pone) node[color=black] {$\bullet$};
            \node at (\i*\incr, {max(\pzero,\pone)+0.08}) {\footnotesize $x^{(\j)}$};
        }
    
        \draw [brace] (3.5*\incr-\gap, -0.2) --  node[below=2pt]{\footnotesize Not targeted} (0.5*\incr+\gap, -0.2);
        \draw [brace] (6.5*\incr-\gap, -0.2) --  node[below=2pt]{\footnotesize Targeted} (3.5*\incr+\gap, -0.2);
        \end{tikzpicture}
        \caption{Ranking of the population of 6 individuals from \cref{ex:profit}. The profit of each individual (vertical axis) is plotted with respect to its score (horizontal axis). Individual action and baseline profits are represented by the full and hollow dots, respectively.}
        \label{fig:example_campaign_profit}
    \end{figure}
    
    In this example, we randomly set the treatment rate to $\rho=0.5$. As such, we obtain a threshold $\tau$ that assigns the treatment $\rv t=1$ to the three individuals with the highest score $\model(x, D_{\mathrm{tr}})$ and assigns $\rv t=0$ to the three remaining individuals. We see from \cref{fig:example_campaign_profit} that the treated individuals are those with features $x^{(2)}, x^{(6)}$ and $x^{(1)}$. The action profit of the campaign (\cref{def:action_profit}) is
    \begin{align*}
        \Pi_1(\rho,D_{\mathrm{tr}})&=\rho\frac 13\left(\pi_1(x^{(2)})+\pi_1(x^{(6)})+\pi_1(x^{(1)})\right)+(1-\rho)\frac 13\left(\pi_0(x^{(5)})+\pi_0(x^{(3)})+\pi_0(x^{(4)})\right)\\
        &=\frac 16(0.05+0.1+0.2+0.2+0.15+0.1)=0.133\dots
    \end{align*}
    Performing this campaign produces an average profit of $0.133\dots$ per individual in the customer base. However, this should be contrasted with the baseline profit (\cref{def:baseline_profit}):
    \begin{equation*}
        \Pi_0 = \frac 16\sum_{i=1}^6 \pi_0(x^{(i)}) = \frac{0.45}6=0.075.
    \end{equation*}
    This represents a significant profit, even though no customers were contacted. The causal profit, in this example is $\Pi(\rho,D_{\mathrm{tr}})=\Pi_1(\rho,D_{\mathrm{tr}})-\Pi_0=0.05833\dots$, indicating that it is still beneficial to perform the campaign. From \cref{thm:total_profit}, we know that this quantity depends only on the causal profit of the targeted individuals, with indices $2, 6$ and $1$ in this example. We can verify this by applying \cref{eq:thm_profit}:
    \begin{align*}
        \Pi(\rho,D_{\mathrm{tr}})&=\frac 16\left(\pi(x^{(2)})+\pi(x^{(6)})+\pi(x^{(1)})\right)=\frac 16(-0.05+0.1+0.3)=0.05833\dots
    \end{align*}
    In this example, the treatment rate $\rho$ can be adjusted to further increase the causal profit. With $\rho=1/3$, only $x^{(1)}$ and $x^{(6)}$ would be targeted, which would result in a causal profit of $0.066\dots$
\end{example}

\subsection{Equivalence with the profit from Verbeke et al.}
\label{sec:equivalence_verbeke}

The profit measure developed in this section is mostly equivalent to the profit measure developed by~\textcite{verbeke2022not} (see \cref{def:verbeke_profit_measure}). The main technical difference is that our measure $\Pi(\rho, D_{\mathrm{tr}})$ naturally accepts individual variations of the cost-benefit matrix $\CB(x)$, i.e., an instance-dependent cost-benefit matrix. Another advantage of our profit measure stems from \cref{thm:total_profit}, which shows that the profit measure can be expressed as $\overline\Pi(\rho)=\E_{\rv x}[\pi(\rv x)P(\model(\rv x, \rv D_{\mathrm{tr}})\ge\tau)]$, highlighting the fact that the profit is determined by two terms, the individual profit $\pi(\rv x)$, which is inherent to the population, and the (stochastic) estimator $\model(\rv x, \rv D_{\mathrm{tr}})$. We discuss this aspect in detail in \cref{sec:influence_profit}. We now show the equivalence with Verbeke's measure.
\repeatedthm{theorem}{verbekeprofit}{thm:verbeke_profit}{
    Let $\CB(x)$ be constant across all $x$, that is, $\CB(x)=\CB$. For any model $\model$ training on a set $D_{\mathrm{tr}}$, and for any threshold $\tau$, we have $\mathrm{CP}(\tau,D_{\mathrm{tr}})=\Pi(\rho,D_{\mathrm{tr}})$ with $\rho=P(\model(\rv x, D_{\mathrm{tr}})\ge\tau)$.
}
This theorem shows that two different definitions of the profit of a campaign lead to the same mathematical quantity. We hope that this firmly establishes the measure of profit as the most general measure of performance for uplift models. Given its generality, in particular stemming from the use of a cost-benefit matrix, the profit measure subsumes most other performance metrics developed in the literature~\parencite{fernandez-loria2022causala,gubela2020response,haupt2022targeting}, and are reviewed in \cref{sec:related_evaluation_measures}.

\subsection{Relationship with the uplift curve}
\label{sec:equivalence_uplift_curve}
The uplift curve, as defined in \cref{def:simple_uplift_curve}, is widely used in the literature to evaluate the performance of uplift models~\parencite{gutierrez2016causal}. While it has been used for more than 20 years~\parencite{lo2002true}, its definition has always been accepted without a formal proof of its validity. Proving this validity requires a definition of the underlying objective of the campaign and a demonstration of the correspondence between this objective and the uplift curve. \textcite{verbeke2021foundations} and this paper propose the causal profit as a definition of the underlying objective of the campaign. In this section, we demonstrate that the uplift curve converges to the measure of profit as the number of samples in the test set increases. Importantly, this convergence is true only under a strong assumption on the cost-benefit matrix, which we call the \emph{unitary value assumption}.
\begin{definition}
    \label{def:unitary_value}
    The \emph{unitary value assumption} posits that the cost-benefit matrix does not depend on $\rv x$ and is    \begin{equation}
        \CB(x)=\begin{bmatrix} 1 & 1 \\ 0 & 0 \end{bmatrix}
    \end{equation}
\end{definition}
Intuitively, the unitary value assumption represents the case where all individuals have an equal value, only the value of the outcome $\rv y$ should be taken into account and the treatment has no cost. These assumptions are rather restrictive, especially the last one. \cref{def:unitary_value} has the following corollary.
\begin{result}
    \label{thm:unitary_value}
    Under the unitary value assumption, the causal profit $\pi(x)$ is equal to the uplift $U(x)$.
\end{result}
\begin{proof}
    From \cref{eq:pi_cb}, by replacing the values of $\CB(x)$, we obtain
    \begin{align*}
        \pi(x)=1-S_1(x)-(1-S_0(x))=S_0(x)-S_1(x)=U(x).
    \end{align*}
\end{proof}
We are now ready to give the main theorem of this section.
\repeatedthm{theorem}{upliftprofit}{thm:uplift_profit}{
    Let $D_{\mathrm{tr}},\rv D_{\mathrm{te}}$ be a training set and a test set of random variables iid to $(\rv x,\rv y,\rv t)$, where $\rv t$ is randomized, and let $\model$ be a model such that $\model(\rv x, D_{\mathrm{tr}})$ is a continuous random variable. Let $N$ be the size of $\rv D_{\mathrm{te}}$, $\rho\in(0,1)$ be the treatment rate, and $k=\ceil{N\rho}$. Under the unitary value assumption, the value of the uplift curve at index $k$, denoted $\mathrm{Uplift}(k,D_{\mathrm{tr}},\rv D_{\mathrm{te}})$, is an estimator of the causal profit of a campaign at the corresponding treatment rate $\rho$. This is expressed formally as
    \begin{equation}
        \lim_{N\rightarrow\infty}\frac 1N\mathrm{Uplift}(k,D_{\mathrm{tr}},\rv D_{\mathrm{te}})= \Pi(\rho,D_{\mathrm{tr}})\quad\text{in probability.}
    \end{equation}
}
This theorem establishes a theoretical foundation for the uplift curve, which is widely used in the uplift literature. It also shows that the unitary value assumption is necessary for the validity of the uplift curve, an assumption that has not been explicitly formulated before. We argue that this assumption, and in particular the implication that the treatment has no cost, might not hold for a large number of practical applications. For example, in churn prediction, different customers represent different values for the company. Marketing efforts should be focused on customers sensitive to the action and who represent a large value. The profit measure takes these two aspects into account, whereas the uplift curve disregards the value of the customer.

\subsection{Empirical profit curve}
\label{sec:empirical_estimator}
In this section, we propose an empirical estimator of the causal profit as a generalization of the uplift curve to an arbitrary cost-benefit matrix. We use a notation similar to that used for the definition of the uplift curve (see \cref{def:simple_uplift_curve}).

\begin{definition}
    \label{def:empirical_profit}
Let $\rv D_{\mathrm{te}}=\{(\rv x^{(i)},\rv y^{(i)},\rv t^{(i)})\}_{i=1}^N$ be a data set of $N$ iid tuples of random variables, such that the treatment $t^{(i)}$ is randomized. Let $\model$ be a model trained on a data set $D_{\mathrm{tr}}$, and let $\rv D_{\mathrm{te}}$ be sorted in decreasing order according to $\model$. For any $i<j$, we have $\model(\rv x^{(i)}, D_{\mathrm{tr}})\ge\model(\rv x^{(j)},D_{\mathrm{tr}})$. The \emph{empirical profit curve} is defined for $k\in\{1,\dots,N\}$ as
\begin{align}
    \label{eq:empirical_profit}
    \widetilde\Pi(k, D_{\mathrm{tr}},\rv D_{\mathrm{te}})=\left(\frac{\tilde{\rv r}_0(k)}{\rv n_0(k)}-\frac{\tilde{\rv r}_1(l)}{\rv n_1(k)}\right)k
\end{align}
with $n_t(k)$ defined as in \cref{def:simple_uplift_curve}, and
\begin{align*}
    \tilde{\rv r}_t(k)=\sum_{i=1}^k\left(\CB_{0t}(x^{(i)})(1-\rv y^{(i)})+\CB_{1t}(x^{(i)})\rv y^{(i)}\right)\I[\rv t^{(i)}=t].
\end{align*}
\end{definition}
It is easy to see that this curve is equivalent to the uplift curve under the unitary value assumption. Indeed, in this case, we have $\CB_{00}^{(i)}=1$ and $\CB_{10}^{(i)}=0$; thus, $\tilde{\rv r}_0(k)=\rv r_0(k)$. Additionally, $\CB_{01}^{(i)}=1$ and $\CB_{11}^{(i)}=0$; thus, $\tilde{\rv r}_1(k)=\rv r_1(k)$. Therefore, \cref{eq:empirical_profit} reduces to \cref{eq:simple_def_uplift_curve}. \cref{thm:uplift_profit} shows the convergence of the uplift curve to the profit measure under the unitary value assumption; however, the proof of the convergence of the empirical profit curve to the empirical profit curve without a unitary value assumption is left for future work.

An important advantage of this empirical estimator is that only the values of the cost-benefit matrix relating to the observed outcome need to be defined. For a given individual with $\rv y^{(i)}=y$ and $\rv t^{(i)}=t$, only the value $\CB_{yt}^{(i)}$ needs to be known. This is especially interesting when the profit in counterfactual scenarios cannot be computed with certainty.

\section{Uplift vs. predictive approach}
\label{sec:uplift_predictive}
In this section, we discuss the different aspects that influence the profit measure, and we apply this discussion to the comparison between the uplift and predictive approaches. In \cref{sec:influence_profit}, we discuss the profit measure in general terms by giving some intuition on its formula. Then, in \cref{sec:sim_norm,sec:sim_dir}, we compare the predictive approaches in two different simulations of increasing complexity. The simpler simulation of \cref{sec:sim_norm} uses a normal distribution for the features and the noise terms, while the more complex simulation of \cref{sec:sim_dir} is based on a Dirichlet distribution, which provides more flexibility and allows the drawing of more general conclusions\footnote{The code of the simulation experiments is available at \url{https://github.com/TheoVerhelst/Uplift-Predictive-Paper}.}.

\subsection{Parameters influencing the profit measure}
\label{sec:influence_profit}
Our profit measure's primary contribution lies in its ability to clearly demonstrate the impact of different components, thereby providing better insights into the performance of a model. Recall from \cref{thm:total_profit} that the profit measure generated by a model $\model$ is expressed as
\begin{equation}
    \label{eq:influence_params}
    \overline\Pi(\rho)=\E_{\rv x}[\pi(\rv x)P(\model(\rv x, \rv D_{\mathrm{tr}})\ge\tau)].
\end{equation}
We discuss each of the two components in the expected value operator in turn.

$\pi(\rv x)$ represents the causal profit of an individual with features $\rv x$. This is specific to the population of the campaign and does not depend on $\model$. Going back to the definition of $\pi(\rv x)$ (\cref{def:ind_causal_profit}), we have
\begin{align*}
    \pi(x)=&\CB_{01}(x)(1-S_1(x))+\CB_{11}(x)S_1(x)-\CB_{00}(x)(1-S_0(x)) - \CB_{10}(x)S_0(x).
\end{align*}
It is clear that the cost-benefit matrix and the distribution of the conditional probabilities $S_0(\rv x)$ and $S_1(\rv x)$ have an influence on the distribution of $\pi(\rv x)$. In particular, when the cost-benefit matrix $\CB(\rv x)$ is constant (i.e., does not depend on $\rv x$), only the conditional probabilities $S_0(\rv x)$ and $S_1(\rv x)$ matter. Recall that the mutual information is a function of the conditional probabilities $S_t(x)=P(\rv y_t=1\mid\rv x=x)$ and the marginal probabilities $S_t=P(\rv y_t=1)$ for $t\in\{0,1\}$~\parencite{cover1991elements}. From this, we can consider three different cases, depending on the mutual information $I(\rv x;\rv y_t)$.
\begin{itemize}
    \item The mutual information $I(\rv x;\rv y_t)$ is the maximum (that is, it is equal to $H(\rv y_t)$). In this case, the scores $S_t(x)$ are either $0$ or $1$, and we can perfectly distinguish between the four counterfactual categories of individuals, which are  persuadable, do-not-disturb, sure thing and lost cause. The optimal targeting is resolved because we can select only persuadable individuals.
    \item The mutual information is zero. In this case, all scores $S_t(x)$ are equal to $S_t=P(\rv y_t=1)$. Since $\CB(x)$ is also constant, the causal profit $\pi(x)$ is the same for all $x$. Any model $\model$ would generate the same benefit as a random model.
    \item The mutual information is between these two extremes. This corresponds to realistic scenarios. The causal profit is influenced by $\CB(x)$ and the scores $S_t(x)$ but also by the prediction model $\model$, as discussed next.
\end{itemize}
In more general settings, where $\CB(\rv x)$ is not the same for all $x$, it is more difficult to draw any conclusion on the distribution of $\pi(\rv x)$ without any other assumption.

The second term, $P(\model(\rv x, \rv D_{\mathrm{tr}})\ge\tau)$, represents the probability that a given individual with features $\rv x$ has a score higher than the threshold $\tau$ with respect to the distribution of training sets $\rv D_{\mathrm{tr}}$. This quantifies the stochastic nature of the learning process. The predicted score varies depending on $\rv D_{\mathrm{tr}}$, and thus, the probability of this score being higher than a threshold varies depending on $\rv D_{\mathrm{tr}}$ as well. To give some intuition, let us suppose that we have two individuals with features $\rv x=x_1$ and $\rv x=x_2$, such that $\pi(x_1)<\pi(x_2)$. A model aiming to rank the most profitable individuals should rank $x_2$ before $x_1$. However, in practice, we only have estimated values $\model(x_1,\rv D_{\mathrm{tr}})$ and $\model(x_2,\rv D_{\mathrm{tr}})$. These estimators can be characterized in terms of their bias and variances. On the one hand, a larger variance always increases the probability of misclassification, i.e., the probability that $\model(x_1,\rv D_{\mathrm{tr}})>\model(x_2,\rv D_{\mathrm{tr}})$. On the other hand, the bias might have a detrimental or a positive effect, depending on its sign. If the bias on $x_2$ is much larger than that on $b_1$, then the profit estimates, although biased, will increase the probability of a correct classification. \textcite{fernandez-loria2022causal} derived an analytical criterion to determine when a model has a higher risk of misclassification than another depending on their biases and variances.

Note, however, that estimating the probability of a correct classification is not the same as estimating the causal profit. This is because the causal profit is determined not only by the two terms in \cref{eq:influence_params} independently but also by their interaction. To demonstrate this, we highlight two important interactions:
\begin{itemize}
    \item Let $\model_1$ and $\model_2$ be two different models that give a high score to different subpopulations, i.e., $P(\model_1(x,\rv D_{\mathrm{tr}})\ge\tau)$ is high only for $x$ in a set $\mathcal X_1$ and $P(\model_2(x,\rv D_{\mathrm{tr}})\ge\tau)$ is high only for $x$ in a different set $\mathcal X_2$. However, if the expected profits $\pi(x)$ on $\mathcal X_1$ and $\mathcal X_2$ are similar, then the causal profit will be similar as well.
    \item Suppose that we have a set of features $\rv x$ that is not very informative. In this case, the posterior probabilities $S_t(\rv x)$ will be close to the prior probability $S_t=\E_{\rv x}[S_t(\rv x)]$. This directly impacts the distribution of $\pi(\rv x)$ as well, such that most individuals have a causal profit $\pi(x)$ close to $\E_{\rv x}[\pi(\rv x)]$ (if $\CB(x)$ does not vary too much across $x$ as well). Since $\pi(x)$, which is estimated by $\model$, does not vary substantially across different values of $\rv x$, then a slight estimation error by $\model$ can easily lead to a misclassification (i.e., ranking the different individuals in the wrong order).
\end{itemize}
These two scenarios show that the causal profit is impacted by the model estimation error and the distribution of $\pi(x)$ in a nontrivial way. In the following sections, we will therefore assess the impact of four components of the problem, which are, the distribution of scores $S_0(x)$ and $S_1(x)$, the cost-benefit matrix $\CB(x)$, the mutual information $I(\rv x; \rv y_t)$ and the estimator variance. In terms of bias, we will focus on the bias inherent to the uplift and predictive approaches. The uplift approach estimates $U(\rv x)$, which is a potentially biased estimator of $\pi(x)$ (if the unitary value assumption does not hold, see \cref{thm:unitary_value}), and the predictive approach estimates $S_0(x)$, which is definitely biased since the impact of $S_1(x)$ is not taken into account.

\subsection{Simulation study with normally distributed features}
\label{sec:sim_norm}
In this section, we demonstrate the influence of mutual information on the performance of the uplift and predictive approaches using a simple data-generating process. Although the results of this simulation might not generalize well to real-life situations, they nonetheless provide qualitative intuitions about the impact of mutual information on the model performance. This also shows that the uplift approach is not always the best option even in very simple settings.

First, we give the mathematical definition of the data-generating process. Let $\rv x$ be a vector of $n$ features $\rv x=[\rv x_1,\dots,\rv x_n]$, which are all independent random variables with a standard normal distribution $\N(0, 1)$.
The binary potential outcomes $\rv y_0$ and $\rv y_1$ are determined using a linear combination of $\rv x$ with coefficient vectors $\lambda_0$ and $\lambda_1\in\mathbb R^n$ and thresholds $\eta_0,\eta_1\in\mathbb R$. More precisely, the outcome $\rv y_t$, for $t=0,1$, is defined as $\rv y_t=\mathbb I[\lambda_t^T\rv x+\rv\varepsilon\ge\eta_t]$ with $\rv\varepsilon\sim\N(0, 1)$.
Higher values of $\eta_t$ lead to a lower probability that $\rv y_t=1$. Additionally, higher values in $\lambda_0$ and $\lambda_1$ induce a lower impact of the noise $\varepsilon$ on the value of $\rv y_t$; hence, the features $\rv x$ are more informative of the outcome $\rv y$ (higher mutual information $I(\rv x; \rv y_t)$). Finally, the treatment indicator $\rv t$ is sampled from a Bernoulli distribution with parameter $p\in(0,1)$, and the observed outcome $\rv y$ is defined accordingly as $\rv y=\rv y_0(1-\rv t) + \rv y_1\rv t$.

With this data-generating process, a training data set $D_{\mathrm{tr}}$ of size $N_{\mathrm{tr}}$ and a test data set $D_{\mathrm{te}}$ of size $N_{\mathrm{te}}$ are generated. The training data set is used to train an uplift model and a predictive model, and their predictions on the test set are then compared in terms of profit measure. In this simulation setting, we have access to the exact conditional probabilities $S_0(x)$ and $S_1(x)$ and to the cost-benefit matrix, and therefore, we know the exact value of the individual profit $\pi(x)$. The conditional probabilities $S_0(x)$ and $S_1(x)$ can easily be retrieved from the distribution of $\rv y_0$ and $\rv y_1$ as follows:
\begin{align*}
    S_t(x)=P(\rv y_t=1\mid\rv x=x)&=P(\lambda_t^Tx+\rv\varepsilon\ge\eta_t)=P(\rv\varepsilon\ge\eta_t-\lambda_t^Tx)\\
    &=1-\Phi(\eta_t-\lambda_t^Tx)=\Phi(\lambda_t^Tx-\eta_t)
\end{align*}
where $\Phi(\cdot)$ is the cumulative distribution function of the standard normal distribution.
To obtain a quantitative measure of performance without relying on the choice of the treatment rate $\rho$, we compute the area under the empirical profit curve:
\begin{equation*}
    \mathrm{AUPC}=\frac 1{N_{\mathrm{te}}}\sum_{k=1}^{N_{\mathrm{te}}}\tilde\Pi(k,D_{\mathrm{tr}},D_{\mathrm{te}}).
\end{equation*}

In this simulation, we also want to assess the impact of the mutual information $I(\rv x; \rv y_0)$ and $I(\rv x; \rv y_1)$ on the performance of the uplift and predictive approaches. The mutual information is computed as $I(\rv x; \rv y_t)=H(\rv y_t)-H(\rv y_t\mid\rv x)$. These two terms are themselves computed as
\begin{align}
    H(\rv y_t)&=-S_t\log S_t-(1-S_t)\log(1-S_t) \label{eq:sim_norm_prior_entropy}\\
    H(\rv y_t\mid\rv x)&=\E_{\rv x}[-S_t(\rv x)\log S_t(\rv x)-(1-S_t(\rv x))\log(1-S_t(\rv x))].\label{eq:sim_norm_post_entropy}
\end{align}
The term $S_t=P(\rv y_t=1)$ in \cref{eq:sim_norm_prior_entropy} is computed as $S_t=\E_{\rv x}[S_t(\rv x)]$. Expected values over the distribution of $\rv x$ (for computing $S_t$ or $H(\rv y_t\mid\rv x)$ in \cref{eq:sim_norm_post_entropy}) are computed by averaging over the sampled data set $D_{\mathrm{te}}$.

The experimental setup is as follows. We use $n=10$ features. The treatment rate is $p=0.04$ to induce a larger variance for the uplift approach. We have $N_{\mathrm{tr}}=1000$ training samples (a low number of samples to induce a high estimator variance) and $N_{\mathrm{te}}=10000$ test samples. The values in the vectors $\lambda_0$ and $\lambda_1$ are chosen randomly according to $\N(1.2c, c^2)$ and $\N(c, c)$, respectively, where $c$ is a scale parameter varying from $10^{-2}$ to $10$ across different runs of the experiment. The thresholds are chosen manually to $\eta_0=1.12$ and $\eta_1=0.87$ to generate a distribution of potential outcomes close to $S_0=0.4$ and $S_1=0.4$. The predictive approach is a logistic regression from the Scikit-learn Python package~\parencite{pedregosa2011scikit} trained on the samples with $\rv t=0$. The uplift approach is a T-learner, implemented as the difference between two logistic regression models (also from Scikit-learn) trained on the control samples ($\rv t=0$) and the target samples ($\rv t=1$), with a regularization parameter $C=10$. We assume a unitary cost-benefit matrix for evaluating the empirical profit curve. The experiment is repeated 100 times, sampling new values for $\lambda_0$ and $\lambda_1$ at each iteration.

\begin{figure}
    \centering
    \includegraphics[width=0.7\linewidth]{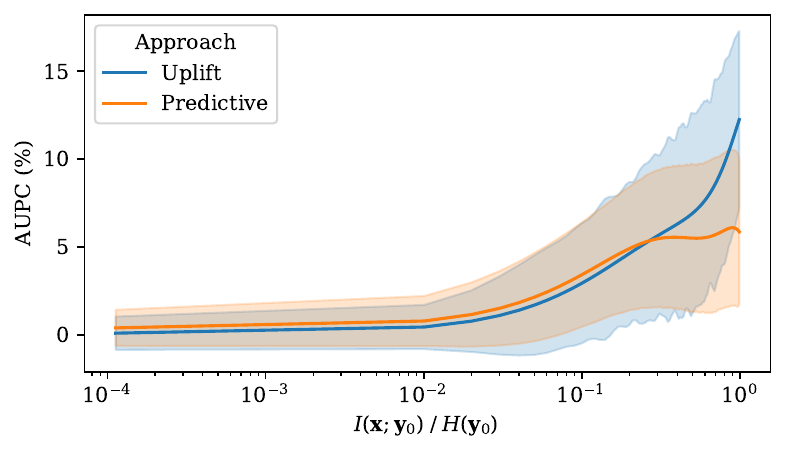}
    \caption{Performance of the uplift and predictive approach as a function of the proportion of mutual information between the features $\rv x$ and $\rv y_0$ in the simulation with normally distributed features. The experiment is repeated 100 times, and the bands represent the 95\% confidence interval. While the uplift approach performs better when all the information is available, the low information regime is dominated by the predictive approach.}
    \label{fig:simulation_norm}
\end{figure}

\cref{fig:simulation_norm} shows the performance of the uplift and predictive approaches as a function of the mutual information $I(\rv x; \rv y_0)$. Note that we report the mutual information $I(\rv x; \rv y_0)$ as a ratio between zero and its maximum value $H(\rv y_0)$. Due to the low number of treated samples in the training set ($p=0.04$ and $N_{\mathrm{tr}}=1000$), the uplift approach suffers from a higher variance than the predictive approach, the latter using only the more numerous control samples. This leads to the predictive approach performing better in terms of the area under the profit curve (AUPC) in the low information regime. However, as the features become more informative, the uplift approach starts to outperform the predictive approach.

This simulated experiment shows that even in very simple settings, with normally distributed features and linear models, the uplift approach does not always provide the best performance. A low number of treated samples and relatively uninformative features can lead the predictive approach to outperform the uplift approach.

\subsection{Simulation study with a Dirichlet distribution}
\label{sec:sim_dir}

In this section, we develop a more complex simulation that allows for more flexibility, and whose characteristics are easier to compute than the simple data-generating process of \cref{sec:sim_norm}. In particular, the distribution of potential outcomes, the mutual information $I(\rv x;\rv y_t)$, and the variance of the uplift and predictive models can all be specified directly and independently of each other. In the simulation from \cref{sec:sim_norm}, these different characteristics had to be computed after the fact because they were influenced by the simulation parameters in a complex way.

In this simulation, we use a sampling process similar to the one used by~\textcite{verhelst2022partial}, in which we do not directly sample the feature distribution $\rv x$. Instead, we directly generate the probabilities of the conditional distribution $\rv y_0,\rv y_1\mid\rv x$. Indeed, the features $\rv x$ influence the result (i.e., the profit measure) only through their impact on the conditional distribution $\rv y_0,\rv y_1\mid\rv x$; hence, if we directly generate the probability of this distribution, then specifying the distribution of $\rv x$ is unnecessary. Afterward, we can emulate the uplift and predictive approaches $\mathcal M_u(x,D_{\mathrm{tr}})$ and $\mathcal M_p(x,D_{\mathrm{tr}})$ as noisy estimates of $U(x)$ and $S_0(x)$ (which are easily computed from the joint distribution of $\rv y_0,\rv y_1\mid\rv x$). Since we sample the distribution $\rv y_0,\rv y_1\mid\rv x$ and the estimated scores but do not sample $\rv x$ directly, we will denote the individual samples with superscript $i$ rather than as functions of $x$.

First, we define additional notation for the joint distribution of the potential outcomes.
\begin{align}
        \alpha &= P(\rv y_0=0, \rv y_1=0) \label{eq:alpha} &
        \beta  &= P(\rv y_0=1, \rv y_1=0) \\
        \gamma &= P(\rv y_0=0, \rv y_1=1) &
        \delta &= P(\rv y_0=1, \rv y_1=1). \label{eq:delta}
\end{align}
We also note $\mu=[\alpha,\beta,\gamma,\delta]$. From this notation, one can easily show that
\begin{align}
    \label{eq:id_S_01}
    S_0=\beta+\delta\quad\text{and}\quad
    S_1=\gamma+\delta.
\end{align}

The sampling process is as follows:
\begin{enumerate}
    \item First, we generate $N$ independent samples $(\rv\alpha^{(i)}, \rv\beta^{(i)}, \rv\gamma^{(i)},\rv\delta^{(i)})_{i=1}^N$ according to a Dirichlet distribution with parameter vector $m=[a,b,c,d]$:
    \begin{align}
        \label{eq:sim_dirichlet}
        (\rv\alpha^{(i)},\dots,\rv\delta^{(i)})\sim\Dir(a,b,c,d).
    \end{align}
    They model the joint probabilities of the potential outcomes as in \crefrange{eq:alpha}{eq:delta} but at the individual level for each individual $i$. The Dirichlet distribution is a natural candidate to sample numbers in a probability simplex (i.e., such that $\rv\alpha^{(i)}, \rv\beta^{(i)}, \rv\gamma^{(i)}$ and $\rv\delta^{(i)}$ are positive and sum up to $1$) since it is the conjugate prior of the multinomial distribution~\parencite{lin2016dirichlet}. This distribution has a number of properties that makes it particularly suited to our setting, which we demonstrate in \cref{sec:appendix_properties_simulation}.

    \item Then, we derive the value of the conditional probabilities $\rv S_0^{(i)}$ and $\rv S_1^{(i)}$ with the identities $\rv S_0^{(i)}=\rv\beta^{(i)}+\rv\delta^{(i)}$ and $\rv S_1^{(i)}=\rv\gamma^{(i)}+\rv\delta^{(i)}$ (\cref{eq:id_S_01}). This results in $\rv S_0^{(i)}$ and $\rv S_1^{(i)}$ having marginal beta distributions, which is convenient because the beta distribution is the conjugate prior of the Bernoulli distribution. This procedure is based on the bivariate beta distribution from~\textcite{olkin2015constructions}.

    \item The score of the predictive approach is emulated as a binomial distribution with parameters $\rv S_0^{(i)}$ and $n_p$ normalized to be at most 1:
    \begin{equation}
        \label{eq:sim_m_p}
        \mathcal M_p^{(i)}\sim\frac 1{n_p}\Bin(\rv S_0^{(i)},n_p).
    \end{equation}
    This emulates the behavior of tree-based models that estimate the conditional probability of the outcome by computing the ratio of positive outcomes among the samples close to the input sample in the feature space. Larger values of $n_p$ induce a lower estimator variance.

    \item Similarly, the score of the uplift approach is emulated as the difference between two normalized binomial distributions, with parameters $\rv S_0^{(i)},n_u$ and $\rv S_1^{(i)},n_u$:
    \begin{equation}
        \mathcal M_u^{(i)}\sim\frac 1{n_u}\Bin(\rv S_0^{(i)},n_u)-\frac 1{n_u}\Bin(\rv S_1^{(i)},n_u).
    \end{equation}

    \item Finally, the binary outcomes $\rv y_0^{(i)},\rv y_1^{(i)}$ are sampled according to a categorical distribution of probability vector $\rv\mu^{(i)}=[\rv\alpha^{(i)},\rv\beta^{(i)},\rv\gamma^{(i)},\rv\delta^{(i)}]$:
    \begin{align}
        P(\rv y_0^{(i)}=0,\rv y_1^{(i)}=0\mid\rv\mu^{(i)}=\mu^{(i)})&=\alpha^{(i)}, \\
        P(\rv y_0^{(i)}=1,\rv y_1^{(i)}=0\mid\rv\mu^{(i)}=\mu^{(i)})&=\beta^{(i)}, \\
        P(\rv y_0^{(i)}=0,\rv y_1^{(i)}=1\mid\rv\mu^{(i)}=\mu^{(i)})&=\gamma^{(i)}, \\
        P(\rv y_0^{(i)}=1,\rv y_1^{(i)}=1\mid\rv\mu^{(i)}=\mu^{(i)})&=\delta^{(i)}.
    \end{align}
\end{enumerate}
This sampling process is particularly convenient because most of its parameters are directly related to the quantities whose impact we wish to assess:
\begin{itemize}
    \item The parameters $a,b,c,d$ are proportional to the probabilities of the distribution of the potential outcomes, $\alpha,\beta,\gamma and \delta$ (see \crefrange{eq:alpha}{eq:delta}). For example, using the moments of the Dirichlet distribution~\parencite{lin2016dirichlet}, we have
    \begin{align*}
        \beta=P(\rv y_0^{(i)}=1,\rv y_1^{(i)}=0)&= \int_{\mathcal S}P(\rv y_0^{(i)}=1,\rv y_1^{(i)}=0\mid\rv\mu^{(i)}=\mu^{(i)})f_{\rv\mu^{(i)}}(\mu^{(i)})\dd\mu^{(i)}\\
        &=\int_{\mathcal S} \beta^{(i)} f_{\rv\mu^{(i)}}(\mu^{(i)})\dd\mu^{(i)}=\E[\rv\beta^{(i)}]=\frac bA
    \end{align*}
    where $A=a+b+c+d$, $\mathcal S$ is the unit 4-simplex, and $f_{\rv\mu^{(i)}}(\cdot)$ is the probability density function of $\rv\mu^{(i)}$.
    \item The conditional entropy $H(\rv y_0^{(i)}, \rv y_1^{(i)}\mid\rv\mu^{(i)})$ is a function of the sum of the parameters $A=a+b+c+d$. The exact analytical relationship between $A$ and the conditional entropy is given in \cref{sec:appendix_properties_simulation}. This conditional entropy is the equivalent in our simulation setting of the entropy conditioned on the features $H(\rv y_0,\rv y_1\mid \rv x)$. As demonstrated in \cref{sec:influence_profit}, this has an important impact on the measure of profit.
    \item The variance of the estimators $\model_u$ and $\model_p$ can be adjusted easily and independently of all the other parameters by choosing the appropriate value of $n_u$ and $n_p$. In fact, we have $\Var(\model_u^{(i)})=\frac 1{n_u}(\rv S_0^{(i)}(1-\rv S_0^{(i)})+\rv S_1^{(i)}(1-\rv S_1^{(i)}))$ and $\Var(\model_p^{(i)})=\frac 1{n_p}(\rv S_0^{(i)}(1-\rv S_0^{(i)}))$. Note that in practice, the variance of the uplift and predictive approaches might not be independent since both are likely to be trained on the same data set.
\end{itemize}

\begin{figure}
    \centering
    \includegraphics[width=0.95\textwidth]{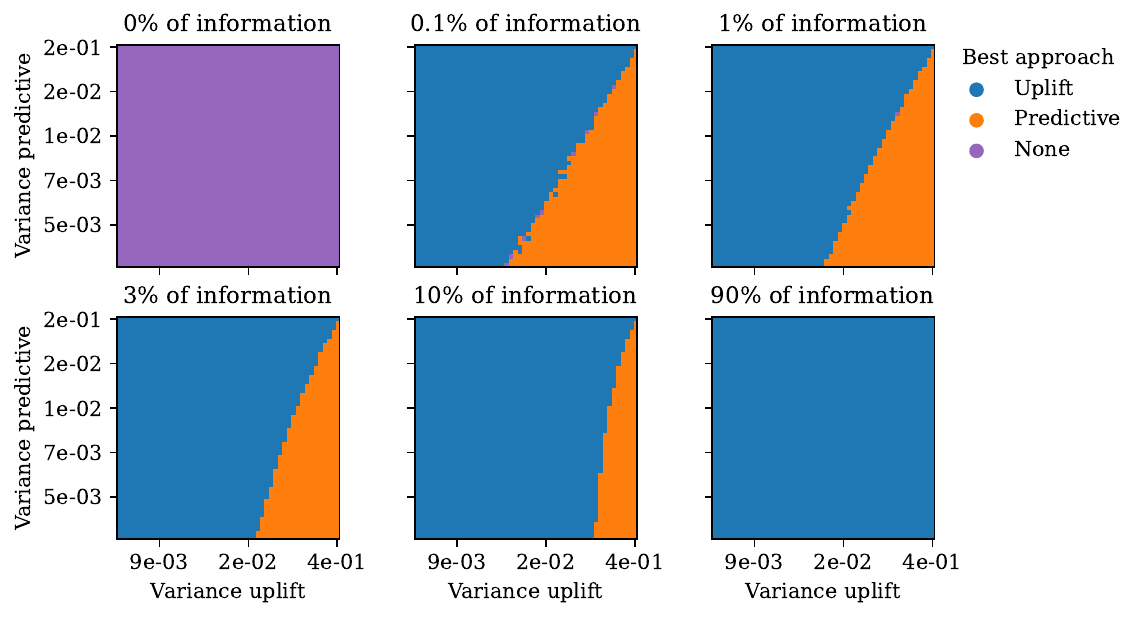}
    \caption{Best approach as a function of the estimator variance for different levels of information between the features and the outcome. Here, $\alpha=0.6,\beta=0.2,\gamma=0.1$ and $\delta=0.1$, and we use the unitary cost-benefit matrix.}
    \label{fig:n_u_n_p_I}
\end{figure}
\begin{figure}
    \centering
    \includegraphics[width=0.95\textwidth]{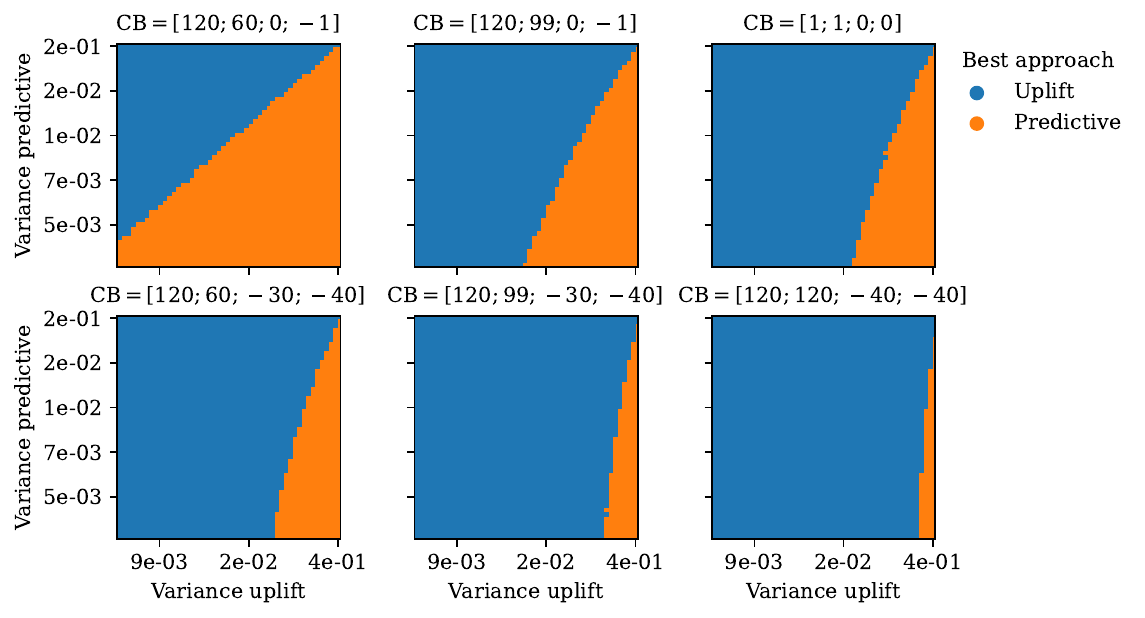}
    \caption{Best approach as a function of the estimator variance for different values of $\CB(x)$. Here, we have $1\%$ of the mutual information between the features and the outcomes. The potential outcome probabilities are $\alpha=0.6,\beta=0.2,\gamma=0.1$ and $\delta=0.1$.}
    \label{fig:n_u_n_p_cb}
\end{figure}
\begin{figure}
    \centering
    \includegraphics[width=0.95\textwidth]{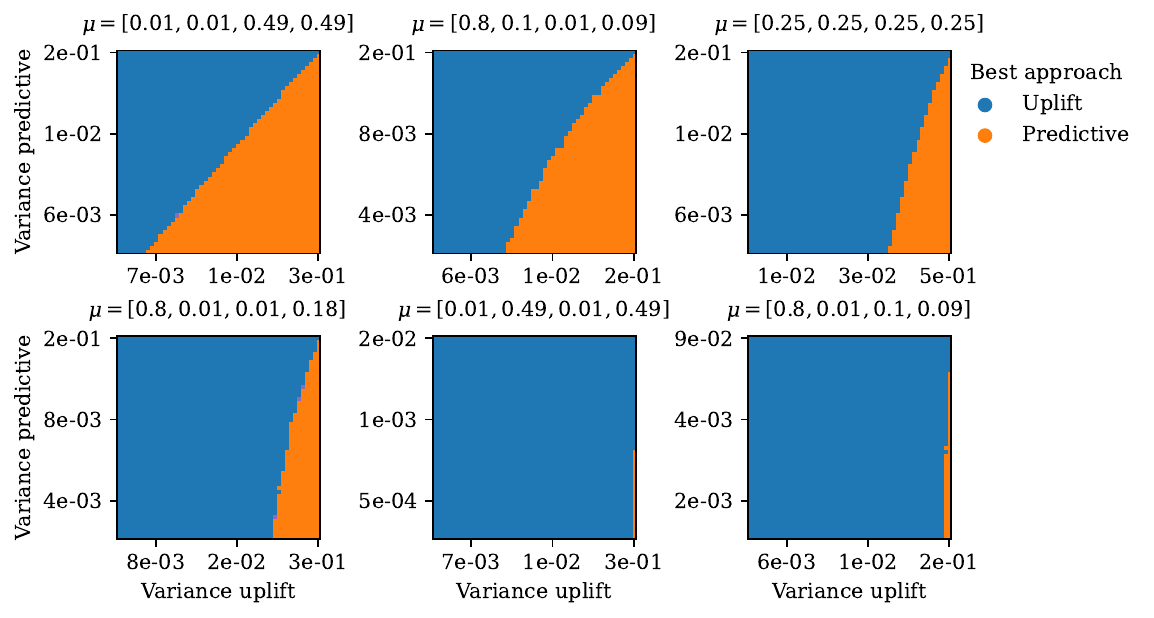}
    \caption{Best approach as a function of the estimator variance for different values of $\mu$. Here, we have $1\%$ of the mutual information between the features and the outcomes.}
    \label{fig:n_u_n_p_mu}
\end{figure}
\begin{figure}
    \centering
    \includegraphics[width=0.55\textwidth]{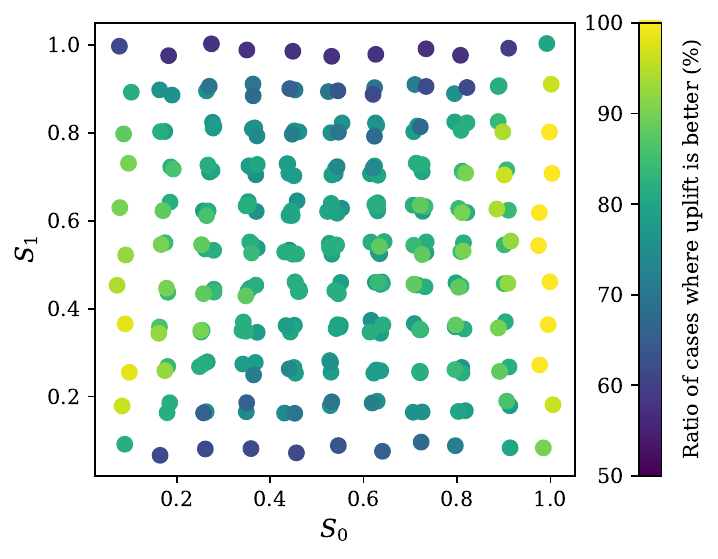}
    \caption{Ratio of runs where the uplift approach is better, where each dot represents several experiments with different estimator variances but a fixed distribution of potential outcomes. The dots are arranged on the space of values of $S_0$ and $S_1$ (remember that $S_0=\beta+\delta$ and $S_1=\gamma+\delta$ from \cref{eq:id_S_01}). Jitter is added to help distinguish overlapping dots.}
    \label{fig:S_0_S_1_uplift_better}
\end{figure}
The sampling process is repeated for different values of the parameters $n_u$ and $n_p$ (both independently from 1 to 50), which influence the variance of the uplift and predictive approaches. \Cref{fig:n_u_n_p_I,fig:n_u_n_p_mu,fig:n_u_n_p_cb} show which of the uplift and predictive approaches performs the best for each value of $n_u$ and $n_p$, and for different values of the other parameters.

\Cref{fig:n_u_n_p_I} varies the mutual information between the features and the outcome. In the first panel, where no information is available, both approaches perform similarly, as they are equivalent to random selection. However, we see on the remaining panels that a lower quantity of information increases the proportion of cases where the predictive approach performs better.

\Cref{fig:n_u_n_p_cb} varies the cost-benefit matrix used to compute the profit measure. The third panel represents the unitary value assumption. The remaining panels highlight various scenarios favoring either the predictive or uplift approach, emphasizing the significant effect of the cost-benefit matrix on their relative performance. We hypothesize that the large difference between, for example, the first and second panels is due to the cost of treatment, i.e., the difference between $\CB_{00}$ and $\CB_{10}$. It is equal to $60$ in the first panel and $21$ in the second panel.

\Cref{fig:n_u_n_p_mu} varies the joint distribution of the potential outcomes across the different panels. We observe that it has an important impact on the performance of the uplift and predictive approaches. In the first panel, where $\gamma=0.49$ and $\delta=0.49$, which indicates a negative causal effect and a high probability $S_1$, the predictive approach performs better than the uplift approach when its variance is lower. On the other hand, in the fifth panel, where $\beta=0.49$ and $\delta=0.49$, the situation is the opposite: the causal effect is high; hence, large benefits can be generated by selecting individuals with the uplift approach in almost every case. The other panels indicate intermediate situations between these two extremes.

To obtain a more comprehensive understanding of the impact of the distribution of potential outcomes, we repeat the experiment for different values of $\mu$ chosen uniformly. Then, for each value of $\mu$, we repeat the experiment by varying the variance of both approaches. \Cref{fig:S_0_S_1_uplift_better} depicts the ratio of cases where the uplift approach outperforms the predictive approach for each value of $\mu$. The data points are arranged according to $S_0$ and $S_1$ (remember that $S_0=\beta+\delta$ and $S_1=\gamma+\delta$ from \cref{eq:id_S_01}). We observe that for a given marginal distribution of the potential outcomes (characterized by $S_0$ and $S_1$), the joint distribution of the potential outcomes (characterized by $\mu$) has a relatively minor impact. Notably, when $S_1$ is close to zero or one while $S_0$ is not, the uplift approach consistently outperforms the predictive approach. Conversely, when $S_0$ is close to zero or one while $S_1$ is not, the predictive approach becomes the favorable choice in approximately 50\% of the cases.

\section{Discussion and limitations}
\label{sec:discussion}
The results presented in this study shed light on the performance of the uplift and predictive approaches under various parameter settings. The findings demonstrate the crucial role played by factors such as the estimator variance, mutual information, cost-benefit matrix and the distribution of the potential outcomes in determining the performance dynamics between these approaches.

One of the key findings of this study is the crucial role played by the variance of the uplift and predictive approaches in determining their relative performance. The results consistently demonstrate that, in almost all settings, one approach will outperform the other if its variance is significantly lower. This highlights the importance of carefully considering the higher variance of the uplift approach compared to the predictive approach.

We summarize our findings as follows. The uplift approach should be preferred when
\begin{itemize}
    \item The outcome is easy to predict from the features, that is, there is a high quantity of  mutual information between the outcome and the features (see \cref{fig:n_u_n_p_I})
    \item The probability of the outcome in the control group ($S_0$) is close to zero or one, while the probability of the outcome in the target group ($S_1$) is not (see \cref{fig:S_0_S_1_uplift_better})
    \item The uplift approach has a lower or the same variance as the predictive approach (see \crefrange{fig:n_u_n_p_I}{fig:n_u_n_p_mu})
\end{itemize}
On the other hand, the predictive approach is more effective when its variance is low enough with respect to the uplift approach, and one of the following conditions is satisfied:
\begin{itemize}
    \item When the mutual information is low (see \cref{fig:n_u_n_p_I})
    \item The probability of the outcome in the target group ($S_1$) is close to zero or one, while the probability of the outcome in the control group ($S_0$) is not (see \cref{fig:S_0_S_1_uplift_better})
    \item The treatment has a significant cost (see \cref{fig:n_u_n_p_cb})
\end{itemize}
The condition of $S_0$ being close to zero or one was already noted in papers comparing the uplift and predictive approaches~\parencite{fernandez-loria2022causal,fernandez-loria2022causala}, but this simulation provides a more systematic illustration. In particular, we observe a symmetric condition where the uplift approach is better when the roles of $S_0$ and $S_1$ are swapped (see \cref{fig:S_0_S_1_uplift_better}).

Our analysis is limited by several factors. First, we assumed that the cost-benefit matrix does not vary across individuals, which could have an important impact on our results. Second, the uplift and predictive approaches have no bias with respect to the quantity they aim to estimate. Machine-learning estimators might be biased, which could further impact their performance. Finally, the choice of modeling the joint distribution of the potential outcomes from a Dirichlet distribution, as well as simulating the estimators from binomial distributions, is obviously a limiting factor. We consider this limitation less critical because the space of distributions that can be simulated in this way is large enough to represent real-world distributions.

\section{Related work}
\label{sec:related_work}
Theoretical analysis of uplift modeling has been the subject of several papers in the literature. In this section, we review the major contributions from the literature on this topic. First, we review the papers proposing new evaluation measures for uplift modeling. Then, we review the few papers explicitly comparing the uplift and predictive approaches.

\subsection{On evaluation measures for uplift modeling}
\label{sec:related_evaluation_measures}

\textcite{fernandez-loria2022causala} discuss the conceptual differences between causal effect estimation and causal classification. In particular, they express the objective of causal classification as the minimization of the expected difference between the best potential outcome and the potential outcome induced by the evaluated model. If we assume that all individuals with a positive uplift should be targeted, then the ideal treatment can be expressed as $\I[U(\rv x)>0]$, and the corresponding potential outcome is noted as $\rv y_{\I[U(\rv x)>0]}$. Similarly, the potential outcome resulting from using a model $\model$ as a decision rule with a threshold of zero is $\rv y_{\I[\model(\rv x, \rv D_{\mathrm{tr}})>0]}$. The evaluation measure proposed by~\textcite{fernandez-loria2022causala} is expressed as
\begin{equation}
    \label{eq:fernandez_regret}
    \mathrm{Regret}(\model)=\E_{\rv x}[\rv y_{\I[\model(\rv x, \rv D_{\mathrm{tr}})>0]} - \rv y_{\I[U(\rv x)>0]}].
\end{equation}
One can see this approach as the converse of our approach and that adopted by~\textcite{verbeke2022not}. Instead of comparing the factual outcome with the outcome resulting from a baseline scenario (such as taking no action), they compare the factual outcome with the best potential outcome that could be taken. However, the baseline profit and the profit generated from the best potential outcomes are independent of the prediction model $\model$; therefore, their values are irrelevant to optimize $\model$. We can show that a model maximizing the regret in \cref{eq:fernandez_regret} also maximizes our measure of profit under the unitary value assumption (this assumption is defined in \cref{def:unitary_value}).

\textcite{li2019unit} define the benefit in terms of the counterfactual category of the customer, which are, persuadable, sure thing, lost cause, or do-not-disturb. They allow for arbitrary costs and benefits for the four different counterfactual categories, and one can view the theoretical framework in this paper, and by~\textcite{verbeke2022not}, as a basis from which one can determine the cost coefficients used by \textcite{li2019unit}. \textcite{zhao2017uplift} generalize the uplift meta-learners to continuous outcomes and multiple treatments. In the evaluation of their methods, they use a metric equivalent to our definition of the action profit (see \cref{def:action_profit}). \textcite{gubela2020response} provide a measure of the profit of a marketing campaign. The formula they propose is tailored to the specific aspects of customer retention (cost of contacting a customer, cost of the incentive, etc.) but does not consider benefits that can vary across individuals.
\textcite{haupt2022targeting} provide a cost-sensitive measure of the profit generated by individuals in the context of customer targeting. They also discuss how to incorporate cost sensitivity into the uplift modeling framework.
Finally, \textcite{gubela2021uplift} propose value-driven evaluation metrics for marketing campaigns, taking into account the trade-off between maximizing uplift and maximizing revenues. Their metric aggregates the estimated uplift and the expected value of individuals into a score used to either rank individuals or evaluate a ranking model. Our work avoids the assumptions underlying the aggregation function and instead evaluates the profit generated by a model based solely on the definition of individual costs and benefits.

\subsection{On uplift vs. predictive approaches}
\label{sec:related_uplift_vs_predictive}
\textcite{devriendt2021why,ascarza2018retention} are the first papers to compare uplift modeling with the predictive approach. Their experimental results suggest that the uplift approach is superior for preventing churn, although their findings are based on a small number of data sets.

\textcite{fernandez-loria2022causal} argue that the true objective is to find \emph{persuadable} individuals, a task named causal classification, and that uplift modeling is only one of the possible ways to tackle causal classification. They derive an analytical criterion expressing when a model outperforms another in terms of classification error, which depends upon the bias and the variance of both models.
Their criterion has two major conceptual differences from the measure of profit developed in our paper. First, it is conditional on a specific value $x$, whereas the profit measure integrates over the probability distribution of $\rv x$. Integrating over the probability distribution of $\rv x$ is essential when the goal is to draw general conclusions about the population rather than about specific individuals. Second, Provost's criterion is based on the probability that a model differs from the Bayes-optimal classifier. However, in practice, an imperfect model might have a large classification error (with respect to the Bayes-optimal classifier) but no loss in terms of profit if the selected individuals have a profit close to that of the individuals selected by the Bayes-optimal classifier.

\textcite{fernandez-loria2022causal,fernandez-loria2022causala} argue that the predictive approach can outperform the uplift approach under four different conditions:
\begin{enumerate}
    \item When positive outcomes are very rare
    \item When the outcomes are difficult to predict
    \item When the treatment effect ($S_0-S_1$) is small
    \item Or when the treatment is correlated with the outcome
\end{enumerate}
Conditions 1 and 4 are generalized in this paper by the result presented in \cref{sec:discussion}, where $S_1$ should be close to zero or one. If $S_1$ is always close to zero or one, then the uplift, which is $S_0-S_1$, will be correlated with the outcome $S_0$. Condition 2 is directly related to the influence of the mutual information discussed in this paper. Condition 3 is partially demonstrated as well in \cref{fig:n_u_n_p_mu}, although we see that the joint distribution of potential outcomes should be considered, not only the treatment effect $S_0-S_1$.
The analysis from Fernández-Loria and Provost is based on the \emph{monotonicity} assumption, i.e., that there is no negative individual causal effect (no \emph{do-not-disturbs}). Additionally, their analysis is based on the area under the ROC curve, which might not always represent the objective of a campaign. We alleviate these two limitations by using the profit measure as the performance metric for comparing the two approaches, and we do not require the monotonicity assumption.

Finally, the work of \textcite{alaa2018limits} investigates the maximum performance of an uplift model using observational data, and provides guidelines to achieve the maximum performance. In the case of experimental data, which is the setting we focus on, their results reduce to the conventional loss of a supervised learning algorithm.

\section{Conclusion}
\label{sec:conclusion}
In this paper, we investigated the effectiveness of uplift modeling compared to the classical predictive approach. To perform this comparison from a sound theoretical basis, we proposed a new formulation of the measure of profit. It emphasizes individual cost sensitivity and the stochastic nature of the underlying machine-learning model. We showed the equivalence of the measure of profit to a preexisting definition in the literature, and the convergence of the uplift curve to the measure of profit. We highlighted the strict conditions necessary for the validity of the uplift curve for a performance evaluation.

The variance of the estimator plays a crucial role in the performance of the uplift and predictive approaches. In almost every case, the predictive approach is preferred if the variance of the uplift approach is high enough. This result is critical, because the uplift approach typically exhibits a higher variance than the predictive approach. The higher variance arises from the fact that the uplift approach estimates the difference between two probabilities, introducing additional uncertainty into the estimation process. We also showed that the mutual information between the input features and the outcome, as well as the distribution of the potential outcomes, plays an important role in determining when the predictive approach outperforms uplift modeling. Last, a proper definition of the cost-benefit matrix is essential, as the performance of both approaches can vary widely depending on it.

Overall, our paper provides firm theoretical foundations for uplift modeling and answers the question of when uplift modeling outperforms predictive modeling. Our findings have important implications for practitioners in various domains, such as marketing, telecommunications, health care and finance, who rely on machine-learning techniques for decision-making. In particular, we suggest that practitioners carefully estimate the various parameters highlighted above to evaluate whether the uplift approach is likely to bring benefits.

In future work, we intend to evaluate our findings on real-world data sets. Preliminary results confirm the importance of mutual information in explaining the differences in the performance of the uplift approach across different data sets. We also intend to assess the consequences of the choice of the uplift learning methodology (S-learner, T-learner, etc.) on the estimator variance in light of these new results.

\section{Acknowledgments}
This work was supported by the Applied PhD Machu-Picchu project funded by Innoviris (2019-PHD-16). We thank this agency for allowing us to conduct both fundamental and applied research.

\appendix

\section{Proofs of equivalence of the profit curve}
\label{sec:appendix_equivalence}
\input{appendix_equivalence}

\section{Properties of the Dirichlet simulation}
\label{sec:appendix_properties_simulation}
\input{appendix_properties_simulation}

\printbibliography

\end{document}

%% file: appendix_equivalence.tex
We first prove the equivalence between Verbeke's measure of profit and the one proposed in this paper.
\repeatverbekeprofit
\begin{proof}
    First, we develop
    \begin{equation*}
        \mathrm{CP}(\tau, D_{\mathrm{tr}})=\mathrm E(\tau, D_{\mathrm{tr}})\oplus\CB=\mathrm{CF}(\tau, D_{\mathrm{tr}})\oplus\CB-\mathrm{CF}(\infty,D_{\mathrm{tr}})\oplus\CB.
    \end{equation*}
    Let's expand the two terms of this difference separately. The first term is
    \begin{align*}
        \mathrm{CF}(\tau, D_{\mathrm{tr}})\oplus\CB=&(1-S_0)F_{00}^{D_{\mathrm{tr}}}(\tau)\CB_{00}+S_0F_{10}^{D_{\mathrm{tr}}}(\tau)\CB_{10}\\
        &+(1-S_1)(1-F_{01}^{D_{\mathrm{tr}}}(\tau))\CB_{01}+S_1(1-F_{11}^{D_{\mathrm{tr}}}(\tau))\CB_{11}.
    \end{align*}
    Let's focus on
    \begin{align*}
        (1-S_0)F_{00}^{D_{\mathrm{tr}}}(\tau)&=(1-S_0)P(\model(\rv x, D_{\mathrm{tr}})<\tau\mid \rv y_0=0)\\
        &=P(\model(\rv x, D_{\mathrm{tr}})<\tau, \rv y_0=0)\\
        &=P(\rv y_0=0\mid \model(\rv x, D_{\mathrm{tr}})<\tau)P(\model(\rv x, D_{\mathrm{tr}})<\tau)\\
        &=P(\rv y_0=0\mid \model(\rv x, D_{\mathrm{tr}})<\tau)(1-\rho)\\
        &=\E[(1-S_0(\rv x))\mid \model(\rv x, D_{\mathrm{tr}})<\tau](1-\rho).
    \end{align*}
    Similarly, we can show that
    \begin{align*}
        S_0F_{10}^{D_{\mathrm{tr}}}(\tau)&=\E[S_0(\rv x)\mid \model(\rv x, D_{\mathrm{tr}})<\tau](1-\rho)\\
        (1-S_1)(1-F_{01}^{D_{\mathrm{tr}}}(\tau))&=\E[(1-S_1(\rv x))\mid \model(\rv x, D_{\mathrm{tr}})\ge\tau]\rho\\
        S_1(1-F_{11}^{D_{\mathrm{tr}}}(\tau))&=\E[S_1(\rv x)\mid \model(\rv x, D_{\mathrm{tr}})\ge\tau)]\rho.
    \end{align*}
    Hence $\mathrm{CF}(\tau, D_{\mathrm{tr}})\oplus\CB$ can be expressed as
    \begin{align*}
        \mathrm{CF}(\tau, D_{\mathrm{tr}})\oplus\CB
        =&\E[(1-S_0(\rv x))\CB_{00}+S_0(\rv x)\CB_{10}\mid \model(\rv x, D_{\mathrm{tr}})<\tau](1-\rho)\\
        &+\E[(1-S_1(\rv x))\CB_{01}+S_1(\rv x)\CB_{11}\mid \model(\rv x, D_{\mathrm{tr}})\ge\tau]\rho\\
        =&\E[\pi_0(\rv x)\mid \model(\rv x, D_{\mathrm{tr}})<\tau](1-\rho)+\E[\pi_1(\rv x)\mid \model(\rv x, D_{\mathrm{tr}})\ge\tau]\rho\\
        =&\Pi_1(\rho, D_{\mathrm{tr}}).
    \end{align*}
    We can use the linearity of the expected value operator to show
    \begin{align*}
        \mathrm{CF}(\infty,D_{\mathrm{tr}})\oplus\CB&=(1-S_0)\CB_{00}+S_0\CB_{10}=(1-\E[S_0(\rv x)])\CB_{00}+\E[S_0(\rv x)]\CB_{10}\\
        &=\E[(1-S_0(\rv x))\CB_{00}+S_0(\rv x)\CB_{00}]=\E[\pi_0(\rv x)]=\Pi_0.
    \end{align*}
    Wrapping up, we have
    \begin{equation*}
        \mathrm{CP}(\tau, D_{\mathrm{tr}})=\mathrm{CF}(\tau, D_{\mathrm{tr}})\oplus\CB-\mathrm{CF}(\infty,D_{\mathrm{tr}})\oplus\CB=\Pi_1(\rho, D_{\mathrm{tr}})-\Pi_0=\Pi(\rho, D_{\mathrm{tr}})
    \end{equation*}
\end{proof}

Now, we prove the equivalence between the profit measure and the uplift curve, stated in \cref{thm:uplift_profit}. Given the length of the proof of \cref{thm:uplift_profit}, it is divided into a series of preliminary results. In these results, we use the notation $\rv q^{(i)}=\rv y^{(i)}\rv t^{(i)}\rv b_k^{(i)}$ and $p=P(\rv t=1)$. Since the training set $D_{\mathrm{tr}}$ is not essential to the proof, we note $\model(\rv x,D_{\mathrm {tr}})=\model(\rv x)$. First, we need another definition of the uplift curve that does not require the test set $\rv D_{\mathrm{te}}$ to be sorted according to the scores $\model(\rv x^{(i)})$. Instead, we define a sequence of binary random variables $(\rv b_k^{(i)})_{i=1}^N$ that indicates which are the $k$ samples from the test set with the highest scores. While our definition is more complex by requiring the definition of $(\rv b_k^{(i)})_{i=1}^N$, it greatly simplifies the proof of \cref{thm:uplift_profit}.
\begin{definition}
    \label{def:uplift_curve}
    Let $\rv D_{\mathrm{te}}=\{(\rv x^{(i)},\rv y^{(i)},\rv t^{(i)})\}_{i=1}^N$ be a data set of $N$ iid tuples of random variables, such that the treatment $t^{(i)}$ is randomized. Let $\model$ be a model, and let $k\in\{1,\dots,N\}$. Let $(\rv b_k^{(i)})_{i=1}^N$ be a sequence of binary random variables indicating the $k$ instances with the highest scores $\model(\rv x^{(i)})$. Formally, $\rv b_k^{(i)}$ is defined such that $\sum_{i=1}^N\rv b_k^{(i)}=k$ and $\model(\rv x^{(i)})\ge \model(\rv x^{(j)})$ whenever $\rv b_k^{(i)}=1$ and $\rv b_k^{(j)}=0$. The \emph{uplift curve} is defined as
    \begin{equation}
        \label{eq:def_uplift_curve}
        \mathrm{Uplift}(k,\rv D_{\mathrm{te}})=\left(\frac{\rv r_0(k)}{\rv n_0(k)}-\frac{\rv r_1(k)}{\rv n_1(k)}\right)k
    \end{equation}
    where the following notation is used, for $t=0,1$:
    \begin{align*}
        \rv r_t(k) &= \sum_{i=1}^N \rv b_k^{(i)}\rv y^{(i)}\I[\rv t^{(i)}=t] & \rv n_t(k) &= \sum_{i=1}^N\rv b_k^{(i)}\I[\rv t^{(i)}=t]
    \end{align*}
    In the case $\rv r_t(k)=\rv n_t(k)=0$, the quotient $\rv r_t(k)/\rv n_t(k)$ is defined as $0$.
\end{definition}
Note that by continuity of $\model(\rv x)$, it is unlikely (i.e., it occurs with probability zero) that multiple instances have the same score from $\model$. Therefore $(\rv b_k^{(i)})_{i=1}^N$ is almost surely uniquely defined. It is easy to see that \cref{def:uplift_curve} is equivalent to \cref{def:simple_uplift_curve}.

\begin{result}
    \label{claim:binom_dis}
    $\rv n_1(k)-1 \mid\rv q^{(i)} = 1$ follows a binomial distribution $\Bin(k-1, p)$.
\end{result}
\begin{proof}
Recall from \cref{def:uplift_curve} that $(\rv b_k^{(i)})_{i=1}^N$ is a sequence of binary random variables defined such that $\sum_i\rv b_k^{(i)}=k$ and $\model(\rv x^{(i)})\ge \model(\rv x^{(j)})$ whenever $\rv b_k^{(i)}=1$ and $\rv b_k^{(j)}=0$. Let $\rv B_k = (\rv b_k^{(i)})_{i=1}^N$. By independence of the $\rv x^{(i)}$'s, $\rv B_k$ is uniformly distributed over the set of sequences with $k$ values equal to one. That is, for any valuation $B_k=(b_k^{(i)})_{i=1}^N$ of $\rv B_k$, one has
\[P(\rv B_k = B_k) =\begin{cases}0 &\text{ if } \sum_i b_k^{(i)} \neq k \\ \binom Nk^{-1} &\text{otherwise.}\end{cases}\]
By independence of the $\rv x^{(i)}$'s, the marginal of each $\rv b_k^{(i)}$ is a Bernoulli trial of probability $k/N$. Let $B_k$ be a realization of $\rv B_k$ such that $\sum_i b_k^{(i)}=k$ and $b_k^{(i)}=1$ for some $i$. Then,
\begin{align*}
    P(\rv B_k=B_k\mid\rv b_k^{(i)}=1) &=\frac{P(\rv b_k^{(i)}=1\mid\rv B_k=B_k)P(\rv B_k=B_k)}{P(\rv b_k^{(i)}=1)}=\frac Nk\binom Nk^{-1}=\binom {N-1}{k-1}^{-1}.
\end{align*}
And that probability is zero if $b_k^{(i)}=0$. Now, note that $\rv n_1(k)\perp\rv y^{(i)}\mid\rv t^{(i)},\rv b_k^{(i)}$, hence $P(\rv n_1(k)=n\mid\rv q^{(i)}=1)=P(\rv n_1(k)=n \mid\rv t^{(i)}\rv b_k^{(i)}=1)$. Also, remember that $\rv n_1(k)=\sum_i\rv t^{(i)}\rv b_k^{(i)}$, hence the expression $\rv n_1(k)=n$ can be decomposed as
\begin{align*}
    P(\rv n_1(k)=n \mid\rv t^{(i)}\rv b_k^{(i)}=1)&=P\left(\rv t_i\rv b_i=1,\sum_{j=1,j\neq i}^N\rv t^{(j)}\rv b_k^{(j)} = n-1 \mid\rv t^{(i)}\rv b_k^{(i)}=1\right)\\
    &=P\left(\sum_{j=1,j\neq i}^N\rv t^{(j)}\rv b_k^{(j)} = n-1 \mid\rv b_k^{(i)}=1\right)
\end{align*}
where the last equality follows from the independence $\rv t^{(i)}\perp\rv t^{(i)}$ for $i\neq j$, and the independence $\rv t^{(i)}\perp\rv b_k^{(j)}$ for any $i,j$ (possibly equal). Marginalizing on all possible values of $\rv B_k$ with $\rv b_k^{(i)}=1$, we have
\begin{align*}
    =&\sum_{\substack{B_k\text{ s.t. }|B_k|=k,\\ b_k^{(i)}=1}}P(\rv B_k=B_k\mid\rv b_k^{(i)}=1)P\left(\sum_{j=1,j\neq i}^N\rv t^{(j)}\rv b_k^{(j)}=n-1\mid\rv B_k=B_k,\rv b_k^{(i)}=1\right) \\
    =&\sum_{\substack{B_k\text{ s.t. }|B_k|=k,\\b_k^{(i)}=1}}\binom {N-1}{k-1}^{-1}P\left(\sum_{j=1,j\neq i}^N\rv t^{(j)}\rv b_k^{(j)}=n-1\mid\rv B_k=B_k\right) \\
    =&\sum_{\substack{B_k\text{ s.t. }|B_k|=k,\\b_k^{(i)}=1}}\binom {N-1}{k-1}^{-1}P(\Bin(k-1, p)=n-1)=P(\Bin(k-1, p)=n-1).
\end{align*}
\end{proof}

\begin{result}
    \label{claim:binom_dis_2}
    $\rv n_1(k)-2\mid\rv q^{(i)}\rv q^{(j)}=1$ follows a binomial distribution $\Bin(k-2,p)$ when $i\neq j$.
\end{result}
\begin{proof}
Let us compute, for any valuation $B_k=(b_k^{(1)},\dots,b_k^{(N)})$ of the random vector $\rv B_k=(\rv b_k^{(1)},\dots,\rv b_k^{(N)})$ and for any $i\neq j$, the probability
\begin{align*}
    P(\rv B_k=B_k\mid\rv b_k^{(i)}\rv b_k^{(j)}=1)&=\frac{P(\rv b_k^{(i)}\rv b_k^{(j)}=1\mid\rv B_k=B_k)P(\rv B_k=B_k)}{P(\rv b_k^{(i)}=1\mid\rv b_k^{(j)}=1)P(\rv b_k^{(j)}=1)}\\
    &=\binom Nk^{-1}\frac{N-1}{k-1}\frac Nk=\binom{N-2}{k-2}^{-1}.
\end{align*}
This allows to develop, for $i\neq j$,
\begin{align*}
    &P\left(\rv n_1(k)=n\mid\rv q^{(i)}\rv q^{(j)}=1\right)\\
    &\qquad=\sum_{\substack{B_k\text{ s.t.}|B_k|=k\\b_k^{(i)}b_k^{(j)}=1}}P\left(\rv n_1(k)=n\mid\rv B_k=B_k,\rv q^{(i)}\rv q^{(j)}=1\right)P(\rv B_k=B_k\mid\rv b_k^{(i)}\rv b_k^{(j)}=1)\\
    &\qquad=\sum_{\substack{B_k\text{ s.t.}|B_k|=k\\b_k^{(i)}b_k^{(j)}=1}}\binom{N-2}{k-2}^{-1}P\left(\sum_l\rv t^{(l)}\rv b_k^{(l)}=n\mid\rv B_k=B_k,\rv q^{(i)}\rv q^{(j)}=1\right)\\
    &\qquad=\sum_{\substack{B_k\text{ s.t.}|B_k|=k\\b_k^{(i)}b_k^{(j)}=1}}\binom{N-2}{k-2}^{-1}P\left(\sum_{l\neq i,j}\rv t^{(l)}\rv b_k^{(l)}=n-2\mid\rv B_k=B_k\right)\\
    &\qquad=\sum_{\substack{B_k\text{ s.t.}|B_k|=k\\b_k^{(i)}b_k^{(j)}=1}}\binom{N-2}{k-2}^{-1}P(\Bin(k-2,p)=n-2)\\
    &\qquad=P(\Bin(k-2,p)=n-2).
\end{align*}
\end{proof}
\begin{result}
    \label{claim:conv_M_b}
    \begin{equation*}
        \lim_{N\to\infty}P(\model(\rv x^{(N)})\ge\tau\mid\rv b_k^{(N)}=1)=\lim_{N\to\infty}P(\rv b_k^{(N)}=1\mid\model(\rv x^{(N)})\ge\tau)=1.
    \end{equation*}
\end{result}
\begin{proof}
This requires a few additional definitions. Let $J=\{j\in\{1,\dots,N\}:\rv b_k^{(j)}=0\}$ be the set of $N-k$ indices with the lowest scores according to $\model$. Since $\model(\rv x)$ is continuous, $J$ is almost surely uniquely defined. Let $\tau_N=\max_{j\in J}\{\model(\rv x^{(j)})\}$ be the threshold that separates the $N-k$ instances with the lowest scores from the $k$ other instances. This is the empirical equivalent of $\tau$ with a dataset of size $N$. We will first prove that $\tau_N$ converges to $\tau$. Remember that $\tau$ is defined as $\tau=\inf\,\{\tau':P(\model(\rv x)\ge\tau')=\rho\}$. Let $F_N$ be the empirical cumulative distribution function of $\{\model(\rv x^{(i)})\}_{i=1}^N$, and let $F_{\model(\rv x)}$ be the cumulative distribution function of $\model(\rv x)$. We have
\begin{align*}
    \lim_{N\to\infty}F_N(\tau_N)&=\lim_{N\to\infty}\frac{N-k+1}N\tag{by def. of $F_N$}\\
    &=1-\rho\tag{by def. of $k$}\\
    &=P(\model(\rv x)<\tau)=F_{\model(\rv x)}(\tau).\tag{by def. of $\tau$}
\end{align*}
The Glivenko-Cantelli theorem states that $\sup_{t\in\mathbb R}|F_N(t)-F_{\model(\rv x)}(t)|$ converges almost surely to zero. By definition of the supremum, we know that, for any $N$, \[|F_N(\tau_N)-F_{\model(\rv x)}(\tau_N)|\le \sup_{t\in\mathbb R}|F_N(t)-F_{\model(\rv x)}(t)|.\] Therefore, $|F_N(\tau_N)-F_{\model(\rv x)}(\tau_N)|$ converges almost surely to zero, and $F_N(\tau_N)-F_{\model(\rv x)}(\tau_N)$ as well. We can now develop
\begin{align*}
    \lim_{N\to\infty}F_N(\tau_N)-F_{\model(\rv x)}(\tau_N)=0\;\Rightarrow\;\lim_{N\to\infty}F_N(\tau_N)&=\lim_{N\to\infty}F_{\model(\rv x)}(\tau_N) =F_{\model(\rv x)}\left(\lim_{N\to\infty}\tau_N\right)
\end{align*}
by continuity of $F_{\model(\rv x)}$. Wrapping up, we have $F_{\model(\rv x)}\left(\lim\tau_N\right)=F_{\model(\rv x)}(\tau)$. If $F_{\model(\rv x)}$ is strictly increasing at $\tau$, then, by continuity, $\lim\tau_N=\tau$. Otherwise, there is a interval $[a,b]$ with $a<b$ such that $F_{\model(\rv x)}(t)=F_{\model(\rv x)}(\tau)$ for any $t\in[a,b]$. Since $F_{\model(\rv x)}\left(\lim\tau_N\right)=F_{\model(\rv x)}(\tau)$, we have $\lim\tau_N\in[a,b]$. Since $F_{\model(\rv x)}(a)=F_{\model(\rv x)}(b)$, we know that $P(a<\model(\rv x)<b)=0$, hence $\lim\tau_N$ is either $a$ or $b$. In fact, it is $a$ because $\tau_N$ is defined as $\max_{j\in J}\{\model(\rv x^{(j)})\}$, the maximum of the $N-k$ lowest scores. Also, by definition of $\tau$, we have $\tau=\inf\,[a,b]=a$. Therefore, $\lim\tau_N=\tau$.

Now, we are ready to prove that $\lim_{N\to\infty}P(\model(\rv x^{(N)})\ge t\mid\rv b_k^{(N)}=1)=1$:
\begin{align*}
    P(\model(\rv x^{(N)})\ge\tau\mid\rv b_k^{(N)}=1)&=P(\model(\rv x^{(N)})\ge\tau\mid \model(\rv x^{(N)})\ge\tau_N)\\
    &=1-P(\model(\rv x^{(N)})<\tau\mid\model(\rv x^{(N)})\ge\tau_N)\\
    &=1-\frac{P(\tau_N\le\model(\rv x^{(N)})<\tau)}{P(\model(\rv x^{(N)})\ge\tau_N)}.
\end{align*}
Since $\tau_N$ converges to $\tau$, $P(\tau_N<\model(\rv x^{(N)})<\tau)$ converges to zero, hence $P(\model(\rv x^{(N)})\ge t\mid\rv b_k^{(N)}=1)$ converges to one. This proves the first part of the claim. We can then use Bayes' theorem to develop
\begin{align*}
    P(\rv b_k^{(N)}=1\mid\model(\rv x^{(N)})\ge t)&=P(\model(\rv x^{(N)})\ge\tau\mid\rv b_k^{(N)}=1)\frac{P(\rv b_k^{(N)}=1)}{P(\model(\rv x^{(N)})\ge\tau)}\\
    &=P(\model(\rv x^{(N)})\ge\tau\mid\rv b_k^{(N)}=1)\frac{k/N}{\rho}
\end{align*}
which converges to the same value as $P(\model(\rv x^{(N)})\ge\tau\mid\rv b_k^{(N)}=1)$, that is, $1$. This proves the second part of \cref{claim:conv_M_b}.
\end{proof}

The following result is purely numerical, and is needed at the end of the proof of \cref{thm:ratio_convergence}.
\begin{lemma}
    \label{claim:convergence_Ak}
    \begin{align*}
        \lim_{k\to\infty}\sum_{n=1}^k\frac 1n\binom knp^n(1-p)^{k-n}=0.
    \end{align*}
\end{lemma}
\begin{proof}
\begin{align*}
    \sum_{n=1}^k\frac 1n\binom knp^n(1-p)^{k-n}&=\sum_{n=1}^k\frac 1n\frac{n+1}{k+1}\binom {k+1}{n+1}p^n(1-p)^{k-n}\\
    &=\frac 1{k+1}\sum_{n=1}^k\frac{n+1}n\binom {k+1}{n+1}p^n(1-p)^{k-n}.
\end{align*}
Since $(n+1)/n\le 2$, we have
\begin{align*}
    \sum_{n=1}^k\frac 1n\binom knp^n(1-p)^{k-n}&\le\frac 2{k+1}\sum_{n=1}^k\binom {k+1}{n+1}p^n(1-p)^{k-n}\\
    &=\frac 2{p(k+1)}\sum_{n=2}^{k+1}\binom{k+1}np^n(1-p)^{k+1-n}\\
    &\le\frac 2{p(k+1)}\sum_{n=0}^{k+1}\binom{k+1}np^n(1-p)^{k+1-n}\\
    &=\frac 2{p(k+1)}\sum_{n=0}^{k+1}P(\Bin(k+1, p)=n)=\frac 2{p(k+1)}.
\end{align*}
Therefore
\begin{align*}
    \lim_{k\to\infty}\sum_{n=1}^k\frac 1n\binom knp^n(1-p)^{k-n}\le\lim_{k\to\infty}\frac 2{p(k+1)}=0.
\end{align*}
Since each term of the sum is positive, the limit is equal to zero.
\end{proof}

The following lemma probably represents the intuition motivating the use of the uplift curve in the literature, although, to the best of our knowledge, it has neither been formalized nor proven before. Its proof is based on all the preceding results.
\begin{lemma}
    \label{thm:ratio_convergence}
    Let $\rv D_{\mathrm{te}}$ be a test set of random variables iid to $(\rv x,\rv y,\rv t)$, where $\rv t$ is randomized, and let $\model$ be a model such that $\model(\rv x)$ is a continuous random variable. Let $N$ be the size of $\rv D_{\mathrm{te}}$, $\rho\in(0,1)$ be the treatment rate, and $k=\ceil{N\rho}$. The ratio $\rv r_1(k)/\rv n_1(k)$, as defined in Definition~\ref{def:uplift_curve}, is a random variable that converges in probability to $P(\rv y=1\mid \model(\rv x)\ge\tau,\rv t=1)$ as $N\rightarrow\infty$, where $\tau=\inf\,\{\tau':P(\model(\rv x)\ge\tau')=\rho)\}$.
\end{lemma}
\begin{proof}
For ease of notation, let $S=P(\rv y=1\mid \model(\rv x)\ge\tau,\rv t=1)$. Also, expressions involving binary variables such as $P(\rv t^{(i)}=1,\rv b_k^{(i)}=1)$ are abbreviated as $P(\rv t^{(i)}\rv b_k^{(i)}=1)$ throughout this proof. The convergence in probability of $\rv r_1(k)/\rv n_1(k)$ to $S$ is formally expressed as, for any $\varepsilon>0$,
\begin{equation}
    \label{eq:proof_conv_def}
    \lim_{N\to\infty}P\left(\left|\frac{\rv r_1(k)}{\rv n_1(k)}-S\right|\ge \varepsilon\right)=0.
\end{equation}
First, we will show the convergence of the expected value of $\rv r_1(k)/\rv n_1(k)$ to $S$:
\begin{equation}
    \label{eq:proof_conv_exp}
    \lim_{N\to\infty}\E\left[\frac{\rv r_1(k)}{\rv n_1(k)}\right]=S.
\end{equation}
From \cref{def:uplift_curve}, we have
\begin{align}
    \label{eq:proof_conv_1}
    \E\left[\frac {\rv r_1(k)}{\rv n_1(k)}\right] &=\E\left[\frac {\sum_{i=1}^N\rv q^{(i)}}{\rv n_1(k)}\right] =\sum_{i=1}^N\E\left[\frac {\rv q^{(i)}}{\rv n_1(k)}\right].
\end{align}
By decomposing the expected value, we see that when $\rv q^{(i)}$ (which is either zero or one) is zero, this specific valuation will not contribute to the expected value. On the other hand, $\rv n_1(k)$ can take any integer value between $0$ and $k$. Hence the expression $\rv q^{(i)}/\rv n_1(k)$ is either zero\footnote{Remember from \cref{def:uplift_curve} that the ratio $\rv r_0(k)/\rv n_0(k)$ is defined as $0$ when $\rv n_0(k)$ equals zero, therefore this case will not contribute to \cref{eq:proof_conv_1}.} or any of the values $1,1/2,\dots,1/k$. This leads to
\begin{align}
    \E\left[\frac {\rv q^{(i)}}{\rv n_1(k)}\right] &=\sum_{n=1}^k\frac 1nP\left(\frac{\rv q^{(i)}}{\rv n_1(k)}=\frac 1n\right)=\sum_{n=1}^k\frac 1nP(\rv n_1(k)=n,\rv q^{(i)}=1) \nonumber\\
    &=\sum_{n=1}^k\frac 1nP(\rv n_1(k)=n \mid\rv q^{(i)}=1)P(\rv q^{(i)}=1).\label{eq:proof_conv_2}
\end{align}
\cref{claim:binom_dis} shows that we can develop \cref{eq:proof_conv_2} as
\begin{align*}
    \E\left[\frac {\rv q^{(i)}}{\rv n_1(k)}\right] &= P(\rv q^{(i)}=1)\sum_{n=1}^k\frac 1nP(\Bin(k-1, p)=n-1).
\end{align*}
This sum can be transformed using the probability mass function of the binomial distribution:
\begin{align*}
    \sum_{n=1}^k\frac 1nP(\Bin(k-1, p)=n-1)
    =&\sum_{n=1}^k\frac 1n\binom {k-1}{n-1}p^{n-1}(1-p)^{k-n}\\
    =&\sum_{n=1}^k\frac 1n\frac {(k-1)!}{(n-1)!(k-n)!}p^{n-1}(1-p)^{k-n} \\
    =&\frac 1{kp}\sum_{n=1}^k\binom knp^n(1-p)^{k-n}=\frac 1{kp}P(\Bin(k,p)\ge 1)\\
    =&\frac 1{kp}(1-P(\Bin(k,p)=0))=\frac 1{kp}\left(1 - (1-p)^k\right).
\end{align*}
Wrapping up, \cref{eq:proof_conv_1} can be developed as
\begin{align*}
    \E\left[\frac {\rv r_1(k)}{\rv n_1(k)}\right] &=\sum_{i=1}^N\E\left[\frac {\rv q^{(i)}}{\rv n_1(k)}\right]=\sum_{i=1}^N\frac 1{kp}P(\rv q^{(i)}=1)\left(1-(1-p)^k\right)\\
    &=\sum_{i=1}^N\frac 1{kp}P(\rv y^{(i)}\rv t^{(i)}\rv b_k^{(i)}=1)\left(1-(1-p)^k\right)
\end{align*}
Since $\rv D_{\mathrm{te}}$ is iid, the choice of the index $i$ in the probability $P(\rv y^{(i)}\rv t^{(i)}\rv b_k^{(i)}=1)$ does not matter. In what follows, we replace it by $N$. Note that $\rv t^{(N)}\perp\rv b_k^{(N)}$, and also $P(\rv b_k^{(N)}=1)=k/N$ and $P(\rv t^{(N)}=1)=P(\rv t=1)=p$, from what we can deduce
\begin{align}
    P(\rv y^{(N)}\rv t^{(N)}\rv b_k^{(N)}=1)&=P(\rv y^{(N)}=1\mid\rv t^{(N)}\rv b_k^{(N)}=1)P(\rv t^{(N)}=1)P(\rv b_k^{(N)}=1)\nonumber\\
    &=\frac{kp}NP(\rv y^{(N)}=1\mid\rv t^{(N)}\rv b_k^{(N)}=1)
    \label{eq:proof_conv_exp_2}
\end{align}
and then
\begin{align*}
    \E\left[\frac {\rv r_1(k)}{\rv n_1(k)}\right]
    &= P(\rv y^{(N)}=1\mid\rv t^{(N)}\rv b_k^{(N)}=1)\left(1-(1-p)^k\right).
\end{align*}
Since $k=\ceil{N\tau}$ and $p\in(0,1)$, it is clear that $\lim_{N\to\infty}(1-(1-p)^k)=1$. To finish the proof of \cref{eq:proof_conv_exp}, it remains to show that
\begin{equation*}
    \lim_{N\to\infty}P(\rv y^{(N)}=1\mid\rv t^{(N)}\rv b_k^{(N)}=1)=S=P(\rv y=1\mid\rv t=1, \model(\rv x) \ge\tau).
\end{equation*}
From \cref{claim:conv_M_b}, we can develop the limit of $P(\rv y^{(N)}=1\mid\rv t^{(N)}\rv b_k^{(N)}=1)$ as
\begin{align*}
    &\lim_{N\to\infty}P(\rv y^{(N)}=1\mid\rv t^{(N)}\rv b_k^{(N)}=1)\\
    &=\lim_{N\to\infty}P(\rv y^{(N)}=1\mid\rv t^{(N)}\rv b_k^{(N)}=1, \model(\rv x^{(N)})\ge\tau)P(\model(\rv x^{(N)})\ge\tau\mid\rv b_k^{(N)}=1)\\
    &\hspace{1em}+\lim_{N\to\infty}P(\rv y^{(N)}=1\mid\rv t^{(N)}b_k^{(N)}=1, \model(\rv x^{(N)})<\tau)P(\model(\rv x^{(N)})<\tau\mid\rv b_k^{(N)}=1)\\
    &=\lim_{N\to\infty}P(\rv y^{(N)}=1\mid\rv t^{(N)}\rv b_k^{(N)}=1, \model(\rv x)\ge\tau).
\end{align*}
In a symmetric fashion, we can develop the limit of $P(\rv y^{(N)}=1\mid\rv t^{(N)}=1,\model(\rv x^{(N)})\ge\tau)$ as
\begin{align*}
    &\lim_{N\to\infty}P(\rv y^{(N)}=1\mid\rv t^{(N)}=1,\model(\rv x^{(N)})\ge\tau)\\
    &=\lim_{N\to\infty}P(\rv y^{(N)}=1\mid\rv t^{(N)}\rv b_k^{(N)}=1,\model(\rv x^{(N)})\ge\tau)P(\rv b_k^{(N)}=1\mid\model(\rv x^{(N)})\ge\tau)\\
    &\hspace{1em}+\lim_{N\to\infty}P(\rv y^{(N)}=1\mid\rv t^{(N)}=1,\rv b_k^{(N)}=0,\model(\rv x^{(N)})\ge\tau)P(\rv b_k^{(N)}=0\mid\model(\rv x^{(N)})\ge\tau)\\
    &=\lim_{N\to\infty} P(\rv y^{(N)}=1\mid\rv t^{(N)}\rv b_k^{(N)}=1,\model(\rv x^{(N)})\ge\tau).
\end{align*}
Using these two convergence results, we have
\begin{align*}
    \lim_{N\to\infty}P(\rv y^{(N)}=1\mid\rv t^{(N)}\rv b_k^{(N)}=1)=&\lim_{N\to\infty}P(\rv y^{(N)}=1\mid\rv t^{(N)}\rv b_k^{(N)}=1,\model(\rv x^{(N)})\ge\tau)\\
    =&\lim_{N\to\infty}P(\rv y^{(N)}=1\mid\rv t^{(N)}=1, \model(\rv x^{(N)}) \ge\tau)\\
    =&P(\rv y=1\mid\rv t=1,\model(\rv x)\ge\tau)
\end{align*}
where the last equality follows by the iid property of $\rv D_{\mathrm{te}}$. This finishes the proof of \cref{eq:proof_conv_exp}:
\begin{align*}
    \lim_{N\to\infty}\E\left[\frac{\rv r_1(k)}{\rv n_1(k)}\right] &= \lim_{N\to\infty}P(\rv y^{(N)}=1\mid\rv t^{(N)}\rv b_k^{(N)}=1)\\
    &= P(\rv y=1\mid\rv t=1,\model(\rv x)\ge\tau)=S.
\end{align*}
Now, let us show that the variance of $\rv r_1(k)/\rv n_1(k)$ converges to $0$:
\begin{equation}
    \label{eq:proof_conv_var}
    \lim_{N\to\infty}\Var\left(\frac{\rv r_1(k)}{\rv n_1(k)}\right)=0.
\end{equation}
We will compute the variance as
\begin{align*}
    \Var\left(\frac{\rv r_1(k)}{\rv n_1(k)}\right)=\E\left[\frac{\rv r_1(k)^2}{\rv n_1(k)^2}\right]-\E\left[\frac{\rv r_1(k)}{\rv n_1(k)}\right]^2.
\end{align*}
Since we just have proven that
$\E[\rv r_1(k)/\rv n_1(k)]$ converges to $S$, we focus on the other term,
$\E[\rv r_1(k)^2/\rv n_1(k)^2]$.
\begin{align}
    \label{eq:proof_conv_var_2}
    \E\left[\frac{\rv r_1(k)^2}{\rv n_1(k)^2}\right]&=\E\left[\frac{\sum_{i=1}^N\rv q^{(i)}\sum_{j=1}^N\rv q^{(j)}}{\rv n_1(k)^2}\right]=\sum_{i=1}^N\sum_{j=1}^N\E\left[\frac{\rv q^{(i)}\rv q^{(j)}}{\rv n_1(k)^2}\right]
\end{align}
Using a similar argument as for the expected value, note that $\rv q^{(i)}\rv q^{(j)}$ is either equal to $0$ or $1$, and that $\rv n_1(k)$ can take any integer value between $0$ and $k$. Hence the expression $\rv q^{(i)}\rv q^{(j)}/\rv n_1(k)^2$ is either zero\footnote{Similarly to \cref{eq:proof_conv_exp}, the expression $\rv q^{(i)}/\rv n_1(k)$ is defined to be zero when $\rv n_1(k)=0$ (in \cref{def:uplift_curve}), therefore the ratio $\rv q^{(i)}\rv q^{(j)}/\rv n_1(k)^2$ is well defined even in that case.} or any of the values $1,1/4,\dots,1/k^2$. This leads to
\begin{align}
    \E\left[\frac{\rv q^{(i)}\rv q^{(j)}}{\rv n_1(k)^2}\right]&=\sum_{n=1}^k\frac 1{n^2}P(\rv n_1(k)=n,\rv q^{(i)}\rv q^{(j)}=1)\nonumber\\
    &=\sum_{n=1}^k\frac 1{n^2}P(\rv n_1(k)=n\mid\rv q^{(i)}\rv q^{(j)}=1)P(\rv q^{(i)}\rv q^{(j)}=1).\label{eq:proof_conv_3}
\end{align}
Using \cref{claim:binom_dis_2}, the sum in \cref{eq:proof_conv_3} can be developed as, for $i\neq j$,
\begin{align*}
    &\sum_{n=1}^k\frac 1{n^2}P(\rv n_1(k)=n\mid\rv q^{(i)}\rv q^{(j)}=1)=\sum_{n=1}^k\frac 1{n^2}P(\Bin(k-2,p)=n-2)\\
    &\qquad=\sum_{n=1}^k\frac {n-1}{n^2(n-1)}\binom{k-2}{n-2}p^{n-2}(1-p)^{k-n}\\
    &\qquad=\frac 1{(k-1)kp^2}\sum_{n=1}^k\frac{n-1}n\binom knp^n(1-p)^{k-n}\\
    &\qquad=\frac 1{(k-1)kp^2}\left(1-(1-p)^k-\sum_{n=1}^k\frac 1n\binom knp^n(1-p)^{k-n}\right).
\end{align*}
Here, we use the notation
\begin{align*}
    A_k&=\sum_{n=1}^k\frac 1n\binom knp^n(1-p)^{k-n}.
\end{align*}
This allows to simplify \cref{eq:proof_conv_3}, for $i\neq j$, as
\begin{align*}
    \E\left[\frac{\rv q^{(i)}\rv q^{(j)}}{\rv n_1(k)^2}\right]&=P(\rv q^{(i)}\rv q^{(j)}=1)\frac{1-(1-p)^k-A_k}{(k-1)kp^2}.
\end{align*}
Now, let us compute, for $i\neq j$, the probability $P(\rv q^{(i)}\rv q^{(j)}=1)$. Remember that $\rv t^{(i)}\perp\rv t^{(j)}$ for $i\neq j$, and that $\rv b_k^{(i)}\perp\rv t^{(j)}$ for any $i,j$ (possibly equal). We also use the fact that $\rv y^{(i)}\perp\rv b_k^{(j)},\rv t^{(j)}\mid\rv b_k^{(i)},\rv t^{(i)}$ for $i\neq j$.
\begin{align}
    P(\rv q^{(i)}\rv q^{(j)}=1)&=P(\rv y^{(i)}\rv t^{(i)}\rv b_k^{(i)}=1\mid\rv y^{(j)}\rv t^{(j)}\rv b_k^{(j)}=1)P(\rv y^{(j)}\rv t^{(j)}\rv b_k^{(j)}=1)\nonumber\\
    &=P(\rv y^{(i)}=1\mid\rv t^{(i)}\rv b_k^{(i)}=1)P(\rv b_k^{(i)}=1\mid\rv b_k^{(j)}=1)P(\rv t^{(i)}=1)\nonumber\\
    &\hspace{4em}P(\rv y^{(j)}=1\mid\rv t^{(j)}\rv b_k^{(j)}=1)P(\rv b_k^{(j)}=1)P(\rv t^{(j)}=1)\nonumber\\
    &=\frac{(k-1)kp^2}{(N-1)N}P(\rv y^{(i)}=1\mid\rv t^{(i)}\rv b_k^{(i)}=1)P(\rv y^{(j)}=1\mid\rv t^{(j)}\rv b_k^{(j)}=1).\label{eq:proof_conv_var_3}
\end{align}
When $i=j$, we have
\begin{align*}
    \E\left[\frac{\rv q^{(i)}\rv q^{(j)}}{\rv n_1(k)^2}\right]&=\sum_{n=1}^k\frac 1{n^2}P(\rv n_1(k)=n\mid\rv q^{(i)}=1)P(\rv q^{(i)}=1)
\end{align*}
From \cref{claim:binom_dis}, the sum can be developed as
\begin{align*}
    \sum_{n=1}^k\frac 1{n^2}P(\rv n_1(k)=n\mid\rv q^{(i)}=1)&=\sum_{n=1}^k\frac 1{n^2}P(\Bin(k-1,p)=n-1)\\
    &=\sum_{n=1}^k\frac 1{n^2}\binom{k-1}{n-1}p^{n-1}(1-p)^{k-n}\\
    &=\frac 1{kp}\sum_{n=1}^k\frac 1n\binom knp^n(1-p)^{k-n}=\frac{A_k}{kp}.
\end{align*}
Hence we have
$\E\left[\rv q^{(i)}\rv q^{(j)}/\rv n_1(k)^2\right]=P(\rv q^{(i)}=1)A_k/(kp)$ for $i=j$. We can develop \cref{eq:proof_conv_var_2} as
\begin{align*}
    &\E\left[\frac{\rv r_1(k)^2}{\rv n_1(k)^2}\right]=\sum_{i=1}^N\sum_{j=1}^N\E\left[\frac{\rv q^{(i)}\rv q^{(j)}}{\rv n_1(k)^2}\right]=\sum_{i=1}^N\sum_{\substack{j=1\\j\neq i}}^N\E\left[\frac{\rv q^{(i)}\rv q^{(j)}}{\rv n_1(k)^2}\right]+\sum_{i=1}^N\E\left[\frac{\rv q^{(i)2}}{\rv n_1(k)^2}\right]\\
    &=\sum_{i=1}^N\sum_{\substack{j=1\\j\neq i}}^N\frac{1-(1-p)^k-A_k}{(k-1)kp^2}P(\rv q^{(i)}\rv q^{(j)}=1)+\sum_{i=1}^N\frac{A_k}{kp}P(\rv q^{(i)}=1).
\end{align*}
Using \cref{eq:proof_conv_exp_2,eq:proof_conv_var_3}, we have
\begin{align*}
    &=\sum_{i=1}^N\sum_{\substack{j=1\\j\neq i}}^N\frac{1-(1-p)^k-A_k}{(k-1)kp^2}\frac{(k-1)kp^2}{(N-1)N}P(\rv y^{(i)}=1\mid\rv t^{(i)}\rv b_k^{(i)}=1)P(\rv y^{(j)}=1\mid\rv t^{(j)}\rv b_k^{(j)}=1)\\
    &\quad+\sum_{i=1}^N\frac{A_k}{kp}\frac{kp}NP(\rv y^{(i)}=1\mid\rv t^{(i)}\rv b_k^{(i)}=1).
\end{align*}
By the iid property of $\rv D_{\mathrm{te}}$, we replace every index $i$ or $j$ by $N$:
\begin{align*}
    &=(1-(1-p)^k-A_k)P(\rv y^{(N)}=1\mid\rv t^{(N)}\rv b_k^{(N)}=1)^2+A_kP(\rv y^{(N)}=1\mid\rv t^{(N)}\rv b_k^{(N)}=1).
\end{align*}
Let $S_N=P(\rv y^{(N)}=1\mid\rv t^{(N)}\rv b_k^{(N)}=1)$. Coming back to \cref{eq:proof_conv_var}, the variance is
\begin{align*}
    \Var\left(\frac{\rv r_1(k)}{\rv n_1(k)}\right)&=\E\left[\frac{\rv r_1(k)^2}{\rv n_1(k)^2}\right]-\E\left[\frac{\rv r_1(k)}{\rv n_1(k)}\right]^2\\
    &=(1-(1-p)^k-A_k)S_N^2+A_kS_N-(1-(1-p)^k)^2S_N^2\\
    &=(1-(1-p)^k-1-(1-p)^{2k}+2(1-p)^k-A_k)S_N^2+A_kS_N\\
    &=((1-p)^k-(1-p)^{2k}-A_k)S_N^2+A_kS_N.
\end{align*}
Its limit is
\begin{align*}
\lim_{N\to\infty}\Var\left(\frac{\rv r_1(k)}{\rv n_1(k)}\right)&=\lim_{N\to\infty}S_N(1-S_N)A_k.
\end{align*}
By \cref{claim:convergence_Ak}, this converges well to zero. 

Using \cref{eq:proof_conv_exp,eq:proof_conv_var}, the cumulative distribution function of $\rv r_1(k)/\rv n_1(k)$, that we note $F^{(N)}$, converges to the unit step function with a step at $S$. That is, $\lim_{N\to\infty}F^{(N)}(x)$ is equal to $0$ when $x<S$ and $1$ when $x\ge S$. Using this, we can develop \cref{eq:proof_conv_def} as
\begin{align*}
    \lim_{N\to\infty}P\left(\left|\frac{\rv r_1(k)}{\rv n_1(k)}-S\right|\ge \varepsilon\right)&=1-\lim_{N\to\infty}P\left(S-\varepsilon<\frac{\rv r_1(k)}{\rv n_1(k)}< S+\varepsilon\right)\\
    &=1-\lim_{N\to\infty}F^{(N)}(S+\varepsilon)+\lim_{N\to\infty}F^{(N)}(S-\varepsilon)\\
    &=1-1+0=0.
\end{align*}
\end{proof}

With \cref{thm:ratio_convergence}, we are now ready to prove \cref{thm:uplift_profit}.
\repeatupliftprofit
\begin{proof}
Let's develop the definition of the uplift curve (\cref{def:simple_uplift_curve}):
\begin{align*}
    \lim_{N\rightarrow\infty}\frac 1N\mathrm{Uplift}(k,D_{\mathrm{tr}},\rv D_{\mathrm{te}})&=\lim_{N\rightarrow\infty}\frac kN\left(\frac{\rv r_0(k)}{\rv n_0(k)}-\frac{\rv r_1(k)}{\rv n_1(k)}\right)\\
    &=\left(\lim_{N\rightarrow\infty}\frac {\ceil{N\rho}}N\right)\left(\lim_{N\rightarrow\infty}\frac{\rv r_0(k)}{\rv n_0(k)}-\lim_{N\rightarrow\infty}\frac{\rv r_1(k)}{\rv n_1(k)}\right).
\end{align*}
It is easy to show that $\lim_{N\to\infty}\ceil{N\rho}/N=\rho$. Using \cref{thm:ratio_convergence} and a similar result for $\rv r_0(k)/\rv n_0(k)$, we have
\begin{align*}
    &\lim_{N\rightarrow\infty}\frac 1N\mathrm{Uplift}(k,D_{\mathrm{tr}},\rv D_{\mathrm{te}})\\
    &=\rho(P(\rv y=1\mid \model(\rv x,D_{\mathrm{tr}})\ge\tau,\rv t=0)-P(\rv y=1\mid \model(\rv x,D_{\mathrm{tr}})\ge\tau,\rv t=1))\\
    &=\rho(P(\rv y_0=1\mid \model(\rv x,D_{\mathrm{tr}})\ge\tau)-P(\rv y_1=1\mid \model(\rv x,D_{\mathrm{tr}})\ge\tau))
\end{align*}
which follows by the randomization of the treatment $\rv t$. Since $\model(\rv x,D_{\mathrm{tr}})$ is a continuous random variance, and by the definition of the threshold $\tau$, we have $\rho=P(\model(\rv x,D_{\mathrm{tr}})\ge\tau)$, leading to
\begin{align*}
    \lim_{N\rightarrow\infty}\frac 1N\mathrm{Uplift}(k,D_{\mathrm{tr}},\rv D_{\mathrm{te}})&=P(\rv y_0=1, \model(\rv x,D_{\mathrm{tr}})\ge\tau)-P(\rv y_1=1, \model(\rv x,D_{\mathrm{tr}})\ge\tau)\\
&=\int f_{\rv x}(x)(P(\rv y_0=1\mid x)-P(\rv y_1=1\mid x))\I[\model(x,D_{\mathrm{tr}})\ge\tau]\dd x\\
&=\int f_{\rv x}(x)U(x)\I[\model(x,D_{\mathrm{tr}})\ge\tau]\dd x\\
&=\E_{\rv x}[U(\rv x)\I[\model(\rv x,D_{\mathrm tr})>\tau]\\
&=\E_{\rv x}[\pi(\rv x)\I[\model(\rv x,D_{\mathrm tr})>\tau]=\Pi(\rho,D_{\mathrm{tr}})
\end{align*}
where the last equalities follows from \cref{thm:unitary_value,thm:total_profit}.
\end{proof}

%% file: appendix_properties_simulation.tex
In this appendix, we provide an analytical formula relating the mutual information and the parameter $A$ in the Dirichlet simulation of \cref{sec:sim_dir}. Note that this is not a closed-form expression, since it is based on the digamma function $\phi$, but it can easily be computed with numerical packages such as SciPy.
\begin{result}
The expected value of the conditional entropy (joint and marginal) of $\rv y_0^{(i)}$ and $\rv y_1^{(i)}$ over the distribution of $\rv\alpha^{(i)},\dots,\rv\delta^{(i)}$ in the simulation of \cref{sec:sim_dir} is
\begin{align}
    \E_{\rv\mu^{(i)}}[H(\rv y_0^{(i)},\rv y_1^{(i)}\mid \rv\mu^{(i)})]&=\psi(A+1)-\sum_{j=1}^4\frac {m_j}A\psi(m_j+1)\\
    \E_{\rv\mu^{(i)}}[H(\rv y_0^{(i)}\mid \rv\mu^{(i)})]&=\psi(A+1)-\left(\frac{b+d}A\psi(b+d+1)+\frac{a+c}A\psi(a+c+1)\right)\\
    \E_{\rv\mu^{(i)}}[H(\rv y_1^{(i)}\mid \rv\mu^{(i)})]&=\psi(A+1)-\left(\frac{c+d}A\psi(c+d+1)+\frac{a+b}A\psi(a+b+1)\right)
\end{align}
where $\psi(x)=\Gamma'(x)/\Gamma(x)$ is the digamma function, and $m=[m_1,m_2,m_3,m_4]=[a,b,c,d]$ is the parameter vector of the Dirichlet distribution in \cref{eq:sim_dirichlet}. Furthermore, as $A\rightarrow0$, the expected conditional entropy (joint and marginal) converges to zero, and, as $A\rightarrow\infty$, the expected conditional entropy (joint or marginal) converges to the corresponding unconditional entropy ($H(\rv y_0^{(i)},\rv y_1^{(i)})$, $H(\rv y_0^{(i)})$ or $H(\rv y_1^{(i)})$).
\end{result}
\begin{proof}
We prove first the derivation of the joint entropy. We have
\begin{equation*}
    H(\rv y_0^{(i)},\rv y_1^{(i)}\mid\rv\mu^{(i)}=\mu)=
    -\alpha\log\alpha-\beta\log\beta-\gamma\log\gamma-\delta\log\delta.
\end{equation*}
We can abbreviate the previous equation as
\begin{equation*}
    H(\rv y_0^{(i)},\rv y_1^{(i)}\mid\rv\mu^{(i)}=\mu)=-\sum_{j=1}^4 \mu_j\log\mu_j.
\end{equation*}
The expected value is
\begin{align*}
    \E_{\rv\mu^{(i)}}[H(\rv y_0^{(i)},\rv y_1^{(i)}\mid\rv\mu^{(i)})]&=
    \int_S H(\rv y_0^{(i)},\rv y_1^{(i)}\mid\rv\mu^{(i)}=\mu)f_{\rv\mu}(\mu)\dd\mu
\end{align*}
where $S$ is the unit 4-simplex $S=\{\mu\mid\mu_j\ge 0\text{ and }\sum_j \mu_j=1\}$, and $f$ is the PDF of the Dirichlet distribution $\Dir(a,b,c,d)$. We can develop the expected value as
\begin{align*}
    \E_{\rv\mu^{(i)}}[H(\rv y_0^{(i)},\rv y_1^{(i)}\mid\rv\mu^{(i)})]&=-
    \int_S \sum_j \mu_j\log\mu_j f_{\rv\mu}(\mu)\dd\mu=-\frac 1 {\B(m)}\sum_j \int_S \mu_j\log\mu_j \prod_k \mu_k^{m_k-1}\dd\mu\\
    &=-\frac 1{\B(m)} \sum_j \mathcal I_j
\end{align*}
where we defined
\begin{align*}
\mathcal I_j&=\int_S \mu_j\log\mu_j \prod_k \mu_k^{m_k-1}\dd\mu= \int_S \mu_j^{m_j}\log\mu_j\prod_{k\neq j}\mu_k^{m_k-1}\dd\mu\\
&=\int_0^1\mu_i^{m_j}\log\mu_j\int_{S(\mu_j)}\prod_{k\neq j}\mu_k^{m_k-1}\dd\mu_{-j}\dd\mu_j
\end{align*}
where we defined $\mu_{-j}=[(\mu_k)]_{k\neq j}$ as the vector of $\mu$ without $\mu_j$, and $S(\mu_j)$ as the domain remaining for $\mu_{-j}$ after $\mu_j$ has taken its value:
\[S(\mu_j)=\Big\{\mu_{-j}:\mu_k>0,\sum_k\mu_k=1-\mu_j\Big\}.\]
We use the Leibniz integral rule and the fact that $\dd a^x/\dd x=a^x\log a$, to express $\mathcal I_j$ as
\begin{align*}
    \mathcal I_j &= \int_0^1\frac{\partial \mu_j^{m_j}}{\partial m_j}\int_{S(\mu_j)}\prod_{k\neq j}\mu_k^{m_k-1}\dd\mu_{-j}\dd\mu_j\\
    &=\frac{\partial}{\partial m_j}\int_0^1\mu_j^{m_j}\int_{S(\mu_j)}\prod_{k\neq j}\mu_k^{m_k-1}\dd\mu_{-j}\dd\mu_j\\
    &=\frac{\partial}{\partial m_j}\int_S\mu_j^{m_j}\prod_{k\neq j}\mu_k^{m_k-1}\dd\mu.
\end{align*}
Note that this last expression is the partial derivative of the Beta function on a vector $m'=[m_1,\dots,m_j+1,\dots,m_4]$. This can also be expressed in terms of the digamma function:
\begin{align*}
    \mathcal I_i&=\frac{\partial\B(m')}{\partial m_j}=\B(m')(\psi(m_i+1)-\psi(A+1))\\
    &=\frac{m_1}{A}\B(m)(\psi(m_i+1)-\psi(A+1))
\end{align*}
where we used the identity
\begin{align*}
    \B(m')=\frac{\Gamma(m_j+1)\prod_{k\neq j}\Gamma(m_k)}{\Gamma(A+1)}=\frac{m_j\Gamma(m_j)\prod_{k\neq j}\Gamma(m_k)}{A\Gamma(A)}=\frac{m_j}{A}\B(m).
\end{align*}
Finally, the expected conditional joint entropy is
\begin{align*}
    \E_{\rv\mu^{(i)}}[H(\rv y_0^{(i)},\rv y_1^{(i)}\mid\rv\mu^{(i)})]&=-\frac 1{\B(m)} \sum_j \mathcal I_j=-\sum_j\frac{m_j}{A}(\psi(m_j+1)-\psi(A+1))\\
    &=\psi(A+1)-\sum_j\frac {m_j}A\psi(m_j+1).
\end{align*}
Note that since the distribution of $\rv y_0^{(i)}$ only depends upon $\rv S_0^{(i)}$, we have that $H(\rv y_0^{(i)}\mid \rv\mu^{(i)})=H(\rv y_0^{(i)}\mid \rv S_0^{(i)})$. Using a similar reasoning as for joint entropy, the marginal entropy is developed as
\begin{align*}
    &\E_{\rv\mu^{(i)}}[H(\rv y_0^{(i)}\mid \rv\mu^{(i)})]=\E_{\rv S_0^{(i)}}[H(\rv y_0^{(i)}\mid \rv S_0^{(i)})]\\
    =&-\int_0^1(s_0\log s_0+(1-s_0)\log(1-s_0))f_{\rv S_0^{(i)}}(s_0)\dd s_0\\
    =&-\frac 1{\B(b+d,a+c)}\int_0^1(s_0\log s_0+(1-s_0)\log(1-s_0))s_0^{b+d-1}(1-s_0)^{a+c-1}\dd s_0\\
    =&-\frac 1{\B(b+d,a+c)}\int_0^1(s_0^{b+d}(1-s_0)^{a+c-1}\log s_0+s_0^{b+d-1}(1-s_0)^{a+c}\log(1-s_0))\dd s_0\\
    =&-\frac 1{\B(b+d,a+c)}\int_0^1\left(\frac{\partial s_0^{b+d}}{\partial(b+d)}(1-s_0)^{a+c-1}+s_0^{b+d-1}\frac{\partial(1-s_0)^{a+c}}{\partial(a+c)}\right)\dd s_0\\
    =&-\frac 1{\B(b+d,a+c)}\left(\frac{\partial\B(b+d+1,a+c)}{\partial(b+d)}+\frac{\partial\B(b+d,a+c+1)}{\partial(a+c)}\right)\\
    =&-\left(\frac{b+d}A(\psi(b+d+1)-\psi(A+1))+\frac{a+c}A(\psi(a+c+1)-\psi(A+1))\right)\\
    =&\psi(A+1)-\left(\frac{b+d}A\psi(b+d+1)+\frac{a+c}A\psi(a+c+1)\right).
\end{align*}
Evidently, a similar development can be found for $\E_{\rv\mu^{(i)}}[H(\rv y_1^{(i)}\mid \rv\mu^{(i)})]$.
\end{proof}

%% file: references.bib
@article{pedregosa2011scikit,
	title = {Scikit-learn: {Machine} {Learning} in {Python}},
	volume = {12},
	issn = {1533-7928},
	shorttitle = {Scikit-learn},
	url = {http://jmlr.org/papers/v12/pedregosa11a.html},
	number = {85},
	urldate = {2023-04-27},
	journal = {Journal of Machine Learning Research},
	author = {Pedregosa, Fabian and Varoquaux, Gaël and Gramfort, Alexandre and Michel, Vincent and Thirion, Bertrand and Grisel, Olivier and Blondel, Mathieu and Prettenhofer, Peter and Weiss, Ron and Dubourg, Vincent and Vanderplas, Jake and Passos, Alexandre and Cournapeau, David and Brucher, Matthieu and Perrot, Matthieu and Duchesnay, Edouard},
	year = {2011},
	pages = {2825--2830},
}

@inproceedings{zhao2017uplift,
	title = {Uplift modeling with multiple treatments and general response types},
	url = {https://epubs.siam.org/doi/abs/10.1137/1.9781611974973.66},
	booktitle = {Proceedings of the 2017 {SIAM} international conference on data mining ({SDM})},
	author = {Zhao, Yan and Fang, Xiao and Simchi-Levi, David},
	year = {2017},
	doi = {10.1137/1.9781611974973.66},
	note = {tex.eprint: https://epubs.siam.org/doi/pdf/10.1137/1.9781611974973.66},
	pages = {588--596},
}

@inproceedings{alaa2018limits,
	series = {Proceedings of machine learning research},
	title = {Limits of estimating heterogeneous treatment effects: {Guidelines} for practical algorithm design},
	volume = {80},
	url = {https://proceedings.mlr.press/v80/alaa18a.html},
	booktitle = {Proceedings of the 35th international conference on machine learning},
	publisher = {PMLR},
	author = {Alaa, Ahmed and van der Schaar, Mihaela},
	editor = {Dy, Jennifer and Krause, Andreas},
	month = jul,
	year = {2018},
	pages = {129--138},
}

@article{holland1986statistics,
	title = {Statistics and {Causal} {Inference}},
	volume = {81},
	issn = {0162-1459, 1537-274X},
	url = {http://www.tandfonline.com/doi/abs/10.1080/01621459.1986.10478354},
	doi = {10.1080/01621459.1986.10478354},
	language = {en},
	number = {396},
	urldate = {2023-06-06},
	journal = {Journal of the American Statistical Association},
	author = {Holland, Paul W.},
	month = dec,
	year = {1986},
	pages = {945--960},
}

@article{haupt2022targeting,
	title = {Targeting customers under response-dependent costs},
	volume = {297},
	number = {1},
	journal = {European Journal of Operational Research},
	author = {Haupt, Johannes and Lessmann, Stefan},
	year = {2022},
	note = {Publisher: Elsevier},
	pages = {369--379},
}

@misc{verhelst2022partial,
	title = {Partial counterfactual identification and uplift modeling: theoretical results and real-world assessment},
	copyright = {Creative Commons Attribution Non Commercial No Derivatives 4.0 International},
	url = {https://arxiv.org/abs/2211.07264},
	publisher = {arXiv},
	author = {Verhelst, Théo and Mercier, Denis and Shrestha, Jeevan and Bontempi, Gianluca},
	year = {2022},
	doi = {10.48550/ARXIV.2211.07264},
}

@article{lin2016dirichlet,
	title = {On the dirichlet distribution},
	journal = {Master's Report},
	author = {Lin, Jiayu},
	year = {2016},
	note = {Publisher: Queen's University Kingston Ontario, Canada},
}

@article{devriendt2021why,
	title = {Why you should stop predicting customer churn and start using uplift models},
	volume = {548},
	journal = {Information Sciences},
	author = {Devriendt, Floris and Berrevoets, Jeroen and Verbeke, Wouter},
	year = {2021},
	note = {Publisher: Elsevier},
	pages = {497--515},
}

@article{gubela2021uplift,
	title = {Uplift modeling with value-driven evaluation metrics},
	journal = {Decision Support Systems},
	author = {Gubela, Robin M and Lessmann, Stefan},
	year = {2021},
	note = {Publisher: Elsevier},
	pages = {113648},
}

@inproceedings{li2019unit,
	title = {Unit {Selection} {Based} on {Counterfactual} {Logic}},
	url = {https://doi.org/10.24963/ijcai.2019/248},
	doi = {10.24963/ijcai.2019/248},
	booktitle = {{IJCAI}},
	publisher = {International Joint Conferences on Artificial Intelligence Organization},
	author = {Li, Ang and Pearl, Judea},
	year = {2019},
	pages = {1793--1799},
}

@article{verbeke2022not,
	title = {To do or not to do? {Cost}-sensitive causal classification with individual treatment effect estimates},
	journal = {European Journal of Operational Research},
	author = {Verbeke, Wouter and Olaya, Diego and Guerry, Marie-Anne and Van Belle, Jente},
	year = {2022},
	note = {Publisher: Elsevier},
}

@article{lo2002true,
	title = {The true lift model: a novel data mining approach to response modeling in database marketing},
	volume = {4},
	number = {2},
	journal = {ACM SIGKDD Explorations Newsletter},
	author = {Lo, Victor S Y},
	year = {2002},
	note = {Publisher: ACM New York, NY, USA},
	pages = {78--86},
}

@article{oskarsdottir2018time,
	title = {Time series for early churn detection: {Using} similarity based classification for dynamic networks},
	volume = {106},
	doi = {10.1016/j.eswa.2018.04.003},
	journal = {Expert Systems with Applications},
	author = {Óskarsdóttir, María and Van Calster, Tine and Baesens, Bart and Lemahieu, Wilfried and Vanthienen, Jan},
	year = {2018},
	note = {Publisher: Elsevier},
	pages = {55--65},
}

@article{verbeke2021foundations,
	title = {The foundations of cost-sensitive causal classification},
	journal = {European Journal of Operational Research},
	author = {Verbeke, Wouter and Olaya, Diego and Berrevoets, Jeroen and Verboven, Sam and Maldonado, Sebastian},
	year = {2021},
	note = {Publisher: Elsevier},
	pages = {1--20},
}

@article{ascarza2018retention,
	title = {Retention futility: {Targeting} high-risk customers might be ineffective},
	volume = {55},
	number = {1},
	journal = {Journal of Marketing Research},
	author = {Ascarza, Eva},
	year = {2018},
	note = {Publisher: SAGE Publications Sage CA: Los Angeles, CA},
	pages = {80--98},
}

@article{gubela2020response,
	title = {Response transformation and profit decomposition for revenue uplift modeling},
	volume = {283},
	number = {2},
	journal = {European Journal of Operational Research},
	author = {Gubela, Robin M and Lessmann, Stefan and Jaroszewicz, Szymon},
	year = {2020},
	note = {Publisher: Elsevier},
	pages = {647--661},
}

@inproceedings{curth2021nonparametric,
	title = {Nonparametric estimation of heterogeneous treatment effects: {From} theory to learning algorithms},
	booktitle = {International {Conference} on {Artificial} {Intelligence} and {Statistics}},
	author = {Curth, Alicia and van der Schaar, Mihaela},
	year = {2021},
	note = {arXiv: Curth2021},
	pages = {1810--1818},
}

@article{mitrovic2018operational,
	title = {On the operational efficiency of different feature types for telco {Churn} prediction},
	volume = {267},
	issn = {03772217},
	url = {https://doi.org/10.1016/j.ejor.2017.12.015},
	doi = {10.1016/j.ejor.2017.12.015},
	number = {3},
	journal = {European Journal of Operational Research},
	author = {Mitrović, Sandra and Baesens, Bart and Lemahieu, Wilfried and De Weerdt, Jochen},
	year = {2018},
	note = {Publisher: Elsevier},
	pages = {1141--1155},
}

@article{kunzel2019metalearners,
	title = {Metalearners for estimating heterogeneous treatment effects using machine learning},
	volume = {116},
	issn = {10916490},
	doi = {10.1073/pnas.1804597116},
	number = {10},
	journal = {Proceedings of the National Academy of Sciences of the United States of America},
	author = {Künzel, Sören R. and Sekhon, Jasjeet S. and Bickel, Peter J. and Yu, Bin},
	year = {2019},
	pmid = {30770453},
	note = {arXiv: 1706.03461
Publisher: National Acad Sciences},
	pages = {4156--4165},
}

@article{devriendt2020learning,
	title = {Learning to rank for uplift modeling},
	journal = {IEEE Transactions on Knowledge and Data Engineering},
	author = {Devriendt, Floris and Van Belle, Jente and Guns, Tias and Verbeke, Wouter},
	year = {2020},
	note = {Publisher: IEEE},
}

@inproceedings{jung2021estimating,
	title = {Estimating {Identifiable} {Causal} {Effects} through {Double} {Machine} {Learning}},
	booktitle = {Proceedings of the 35th {AAAI} {Conference} on {Artificial} {Intelligence}},
	author = {Jung, Yonghan and Tian, Jin and Bareinboim, Elias},
	year = {2021},
}

@inproceedings{idris2014ensemble,
	title = {Ensemble based efficient churn prediction model for telecom},
	doi = {10.1109/fit.2014.52},
	booktitle = {Frontiers of {Information} {Technology} ({FIT}), 2014 12th {International} {Conference} on},
	author = {Idris, Adnan and Khan, Asifullah},
	year = {2014},
	pages = {238--244},
}

@book{cover1991elements,
	title = {Elements of information theory},
	isbn = {0-471-06259-6},
	publisher = {John Wiley \& Sons},
	author = {Cover, Thomas M. and Thomas, Joy A.},
	year = {1991},
	doi = {10.1002/0471200611},
	note = {Publication Title: Elements of Information Theory},
}

@article{olkin2015constructions,
	title = {Constructions for a bivariate beta distribution},
	volume = {96},
	journal = {Statistics \& Probability Letters},
	author = {Olkin, Ingram and Trikalinos, Thomas A},
	year = {2015},
	note = {Publisher: Elsevier},
	pages = {54--60},
}

@book{pearl2009causality,
	title = {Causality: models, reasoning, and inference},
	isbn = {978-0-521-89560-6},
	publisher = {Cambridge university press},
	author = {Pearl, Judea},
	year = {2009},
}

@article{fernandez-loria2022causal,
	title = {Causal {Classification}: {Treatment} {Effect} {Estimation} vs. {Outcome} {Prediction}},
	volume = {23},
	number = {59},
	journal = {Journal of Machine Learning Research},
	author = {Fernández-Loria, Carlos and Provost, Foster},
	year = {2022},
	pages = {1--35},
}

@article{fernandez-loria2022causala,
	title = {Causal decision making and causal effect estimation are not the same… and why it matters},
	journal = {INFORMS Journal on Data Science},
	author = {Fernández-Loria, Carlos and Provost, Foster},
	year = {2022},
	note = {Publisher: INFORMS},
}

@inproceedings{gutierrez2016causal,
	address = {Microsoft NERD, Boston, USA},
	title = {Causal {Inference} and {Uplift} {Modelling}: {A} {Review} of the {Literature}},
	volume = {67},
	url = {http://proceedings.mlr.press/v67/gutierrez17a.html},
	booktitle = {Proceedings of {The} 3rd {International} {Conference} on {Predictive} {Applications} and {APIs}},
	publisher = {PMLR},
	author = {Gutierrez, Pierre and Gérardy, Jean-Yves},
	editor = {Hardgrove, Claire and Dorard, Louis and Thompson, Keiran and Douetteau, Florian},
	month = jan,
	year = {2016},
	note = {Series Title: Proceedings of Machine Learning Research},
	pages = {1--13},
}

@article{zhu2017empirical,
	title = {An empirical comparison of techniques for the class imbalance problem in churn prediction},
	volume = {408},
	doi = {10.1016/j.ins.2017.04.015},
	journal = {Information sciences},
	author = {Zhu, Bing and Baesens, Bart and vanden Broucke, Seppe K L M},
	year = {2017},
	note = {Publisher: Elsevier},
	pages = {84--99},
}

@article{zhang2021unified,
	title = {A unified survey of treatment effect heterogeneity modelling and uplift modelling},
	volume = {54},
	number = {8},
	journal = {ACM Computing Surveys (CSUR)},
	author = {Zhang, Weijia and Li, Jiuyong and Liu, Lin},
	year = {2021},
	note = {Publisher: ACM New York, NY},
	pages = {1--36},
}
